\documentclass[pdflatex,sn-nature]{sn-jnl}

\usepackage{graphicx}%
\usepackage{multirow}%
\usepackage{amsmath,amssymb,amsfonts}%
\usepackage{amsthm}%
\usepackage{mathrsfs}%
\usepackage[title]{appendix}%
\usepackage{xcolor}%
\usepackage{textcomp}%
\usepackage{manyfoot}%
\usepackage{booktabs}%
\usepackage{algorithm}%
\usepackage{algorithmicx}%
\usepackage{algpseudocode}%
\usepackage{listings}%
\usepackage{natbib}
\usepackage{pifont}
\newcommand{\xmark}{\ding{55}}%
\usepackage{tabularx}

\theoremstyle{thmstyleone}%
\theoremstyle{thmstyletwo}%

\theoremstyle{thmstylethree}%

\begin{document}

\title[Article Title]{Zero-Shot Personalized Camera Motion Control for Image-to-Video Synthesis}


\author[1]{\fnm{Pooja} \sur{Guhan}}\email{pguhan@umd.edu}

\author[1]{\fnm{Divya} \sur{Kothandaraman}}\email{dkr@umd.edu}

\author[1]{\fnm{Geonsun} \sur{Lee}}\email{gsunlee@umd.edu}
\author[2]{\fnm{Tsung-Wei} \sur{Huang}}\email{tsung-wei.huang@dolby.com}
\author[2]{\fnm{Guan-Ming} \sur{Su}}\email{guanming.su@dolby.com}
\author[1]{\fnm{Dinesh} \sur{Manocha}}\email{dmanocha@umd.edu}

\affil[1]{\orgname{University of Maryland College Park}, \orgaddress{\state{Maryland}, \country{USA}, \postcode{20742},}}

\affil[2]{\orgname{Dolby Laboratories}, \orgaddress{\city{Sunnyvale}, \state{California}, \country{USA}, \postcode{94085}}}


\abstract{Specifying nuanced and compelling camera motion remains a significant hurdle for non-expert creators using generative tools, creating an “expressive gap” where generic text prompts fail to capture cinematic vision. This barrier limits individual creativity and restricts the accessibility of cinematic production for small-scale industries and educational content creators. To address this, we present a zero-shot diffusion-based framework for personalized camera motion control, enabling the transfer of cinematic movements from a single reference video onto a user-provided static image without requiring 3D data, predefined trajectories, or complex graphical interfaces. Our technical contribution involves an inference-time optimization strategy using dual Low-Rank Adaptation (LoRA) networks, with an orthogonality regularizer that encourages separation between spatial appearance and temporal motion updates, alongside a homography-based refinement strategy that provides weak geometric guidance. We evaluate our approach using a new metric, CameraScore, and two distinct user studies. A 72-participant perceptual study demonstrates that our method significantly outperforms existing baselines in motion accuracy (90.45\% preference) and scene preservation (70.31\% preference). Furthermore, a 12-participant task-based interaction study confirms that our workflow significantly improves usability and creative control (p $<$ 0.001) compared to standard text- or preset-based prompts. We hope this work lays a foundation for future advancements in camera motion transfer across diverse scenes.}




\maketitle

\begin{figure}
    \centering
    \includegraphics[width=\textwidth]{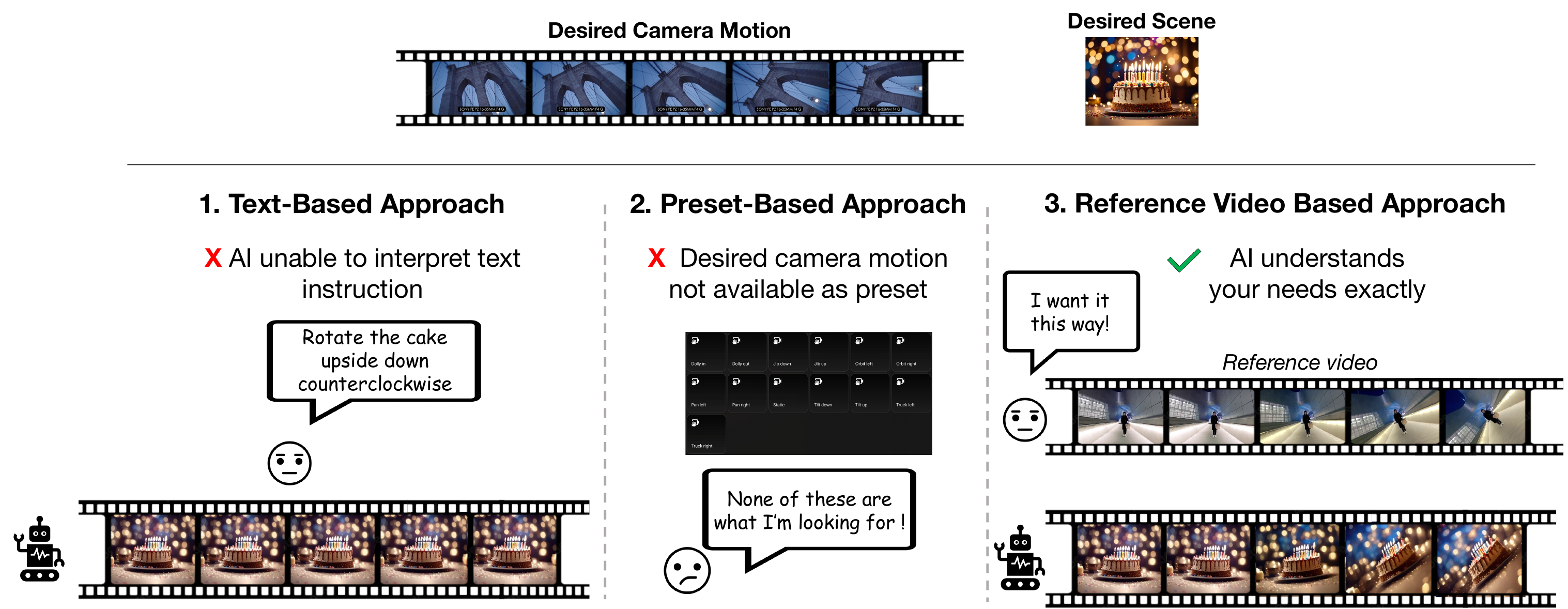}
    \caption{A comparison of three types of instruction interface available to replicate a desired camera motion on a desired scene. (1) Text Based Approach: The instruction may not be interpreted by AI correctly (2) Preset Based Approach: The preset with the exact camera motion may not be available for AI to use (3) Reference Video Based Approach: Intuitive input and AI understands what you need exactly.}
    \label{fig:cover}
\end{figure}
\section{Introduction}
\label{sec:intro}

Digital video has become a vital medium for communication and creative expression. The rise of generative artificial intelligence (AI)~\cite{van2013generative}~offers unprecedented opportunities for both professional and novice creators, enabling the synthesis of personalized content with unique characters~\cite{zhang2024pia}, customized backgrounds~\cite{li2025tuning}, and tailored motions~\cite{wu2024motionbooth, ma2024follow}. Recent advances in personalized image~\cite{shi2024instantbooth, li2024photomaker, he2024imagine} and video generation~\cite{guo2023animatediff, xing2024dynamicrafter,ma2025video} empower creators to infuse their work with distinctive elements, pushing the boundaries of creativity well beyond training data limitations and finding significant application in animation and filmmaking~\cite{totlani2023evolution}.

Yet, a critical component of cinematic narrative remains elusive: \textit{camera motion}~\cite{ref1}. This \textit{subtle but impactful visual shifting and movement guides perspective}, evokes emotion~\cite{cameramotiondefn, videographertips, rubberducksmovementstory}, and often determines the difference between a static scene and a captivating sequence. Current generative systems offer only limited control. Abstract text prompts (e.g., “zoom in”) lack the granularity for nuanced cinematic intent, while graphical motion panels require technical skill and patience that reinforces the expressive gap for casual creators. This barrier not only limits individual creativity but also restricts the accessibility of cinematic production for small-scale industries and educational content creators, aligning with broader concerns about inclusive innovation highlighted in UN Sustainable Development Goal 9~\cite{assembly2015sustainable}.

To bridge this gap, we introduce \textit{exemplar-based camera motion personalization}: the ability to transfer the specific cinematic movements observed in a reference video onto a new scene of interest. In this paradigm, users can \textit{show and tell} the desired motion by supplying a short clip whose camera behavior they like, and applying that motion to their own still image. This capability democratizes video production~\cite{productionstory}, allowing creators to borrow and adapt professional techniques effortlessly. However, existing methods fall short. Video foundation models like MovieGen~\cite{moviegen_meta} and Veo~\cite{veo2} produce generic motion effects tied to text prompts, not personalized trajectories. Related motion transfer works, such as COMD~\cite{hu2024comd} and AnimateDiff~\cite{guo2023animatediff}, rely on multiple reference videos, which is poorly suited for capturing the distinctive pacing and angles of a single exemplar.
While recent foundation models and motion transfer methods enable impressive video synthesis, they either expose only coarse text-level control or require multiple references, explicit 3D trajectories, or specialized motion modules. We instead focus on the practical scenario of \textit{single-exemplar, cross-scene camera motion transfer} onto \textit{arbitrary} static images using off-the-shelf diffusion backbones.

Controllable camera motion is central not only to professional cinematography but also to emerging AI-mediated media workflows, from short-form online video to virtual production and immersive experiences. By lowering the technical barrier to reusing cinematic camera styles, our method illustrates how AI for video processing can support creative industries, democratize access to advanced visual storytelling techniques, and contribute to innovation in digital content production.

\paragraph{Main Contributions}
We present a method for \textit{personalized camera motion transfer} from a \textit{single} reference video to a static image, empowering creators to realize their own scenes with cinematic movements directly inspired by professional footage. Our contributions are summarized as follows:
\begin{enumerate}
\item We introduce a zero-shot approach that \textit{transfers camera motion from a single reference video to a static image}, without requiring additional camera trajectories or 3D reconstruction. Our key technical component is an inference-time optimization on a pretrained text-to-video diffusion model using dual LoRA networks with an orthogonality regularizer that encourages separation between spatial appearance updates and temporal motion updates. We further leverage homography representations from classical computer vision to guide the generated video along the desired camera motion, producing coherent and realistic sequences.
\item For quantitative evaluation of this task, we introduce CameraScore, a homography-based metric that assesses temporal consistency between the dominant camera-induced motion in the generated video and in the reference, and we report it alongside established representation and text–video alignment metrics. Extensive quantitative and qualitative analyses, including a comparison study with 72 participants, demonstrate that our method significantly outperforms relevant prior methods, achieving 90.45\% preference for motion accuracy and 70.31\% preference for scene preservation.
\item We design an intuitive workflow that allows creators to animate still images simply by providing a reference video, avoiding reliance on abstract prompts or generic presets. A user study with 12 participants confirms that our interface improves usability and creative control significantly compared to standard text- or preset-based prompts (p$<$0.001).
\end{enumerate}

\section{Related Works} \label{sec:our_approach}
AI techniques have been widely applied to video and motion processing tasks, including reconstructing sharp video bursts from a single motion-blurred image~\cite{yosef2023video} and stereoscopic video deblurring with transformer architectures~\cite{imani2024stereoscopic}. These works illustrate the broader role of AI in understanding and manipulating motion in video. In this paper, we focus instead on controllable camera motion transfer and creator-facing interaction, rather than recovering latent motion from degraded footage.

\subsection{User Interfaces for Video Creation and Controllability}
Traditional video editing tools pose significant barriers due to complexity, driving research toward accessible interfaces leveraging natural language and sketching~\cite{tilekbay2024expressedit}. Modern generative models offer powerful text-to-video synthesis~\cite{veo2, photoshop2021adobe}, but \textit{lack geometric control}. Integrating cinematic camera motion \textit{requires specialized expertise} for manual calibration, homography estimation, and 3D trajectory reconstruction—a major hurdle for non-specialists~\cite{abbaspour2017practical}. 
While recent works such as VGGT~\cite{wang2025vggt}~provide strong automatic estimators of camera motion, depth and point clouds from monocular videos, they still require non-trivial integration and may not cover the full diversity of the in-the-wild, stylized or low-light footage creators wish to repurpose. Our approach takes a \textit{complementary} stance. Instead of predicting full 3D trajectories, we use homography-based 2D cues for weak guidance that can be obtained directly from the reference video and seamlessly plugged into existing diffusion pipeline. 

\subsection{Video Personalization, Controllability, and Motion Transfer}

Advancements in pretrained models~\cite{blattmann2023stable, wang2023modelscope, polyak2024movie} have fueled interest in fine-grained control~\cite{jeong2024vmc, ren2024customize, zhang2023motioncrafter, materzynska2023customizing, wang2024customvideo, kothandaraman2024text, guo2023animatediff}.

\textit{Fine-tuning and Local Motion:} Zero-shot methods like Tune-A-Video~\cite{wu2023tune, ren2024customize} use spatial/temporal LoRA~\cite{hu2021lora} to personalize based on a few images/videos. While effective for local subject motion, they struggle with camera motion transfer due to the difficulty in separating scene characteristics from camera dynamics during fine-tuning.

\textit{Explicit Control \& Trajectories:} Other methods use specialized motion modules~\cite{hu2024motionmaster, guo2023animatediff} or require explicit camera inputs such as 3D trajectories and SfM poses~\cite{he2024cameractrl, zhang2024recapture,yu2025trajectorycrafter} or detailed natural language descriptions of camera behavior~\cite{jiang2024cinematographic}. Approaches based on 3D/4D reconstruction, including ReCapture~\cite{zhang2024recapture}, also incur substantial computational and data costs. In a related direction, CamCloneMaster~\cite{luo2025camclonemaster}, ReCamMaster~\cite{bai2025recammaster}, and Akira~\cite{wang2025akira}~explore camera control or cinematic transfer using multiple reference videos or NeRF/ray-based scene representations, and therefore assume access to accurate camera poses or structured scene geometry. By contrast, we focus on single-exemplar, cross-scene camera motion transfer onto arbitrary static images using only 2D cues derived from the reference video and off-the-shelf text-to-video diffusion backbones, without requiring any additional 3D reconstruction or motion annotations.

\subsection{Image Personalization and Novel View Synthesis (NVS)}

Image personalization excels at customizing objects, poses, and styles~\cite{chen2024anydoor, yuan2023customnet, ma2017pose, nguyen2024nope, xu2024sgdm, ruiz2024hyperdreambooth, tewel2023key, li2025tuning, alaluf2023neural}. However, applying these to video motion is challenging due to the missing temporal dimension. Similarly, methods like ControlNet~\cite{zhang2023adding} require substantial supervised fine-tuning.

Our task, transferring camera motion to a static image, is akin to Novel View Synthesis (NVS)~\cite{shi2023zero123++, liu2024one, zhang2024recapture}, which generates frames from new perspectives. A key limitation of NVS, however, is the typical requirement for detailed camera trajectory information to define the desired view angle. Robust 4D NVS also often struggles with generalizability and requires extensive training data. Our objective is unique: to develop a \textit{generalizable, data-efficient, zero-shot} approach that implicitly transfers camera motions from a single reference video, thereby circumventing the need for explicit trajectory input or costly domain-specific data collection.

\section{Method}
\label{sec:our_approach}

To enable personalized and intuitive camera motion transfer (CMT), our system utilizes a pretrained text-to-video diffusion model conditioned on visual content and structured text prompts. This foundational method powers our zero-shot transfer pipeline, integrating complex camera motion from a single reference video onto a user-provided image without requiring explicit 3D reconstruction.

\begin{figure*}
    \centering
    \includegraphics[width=0.9\textwidth]{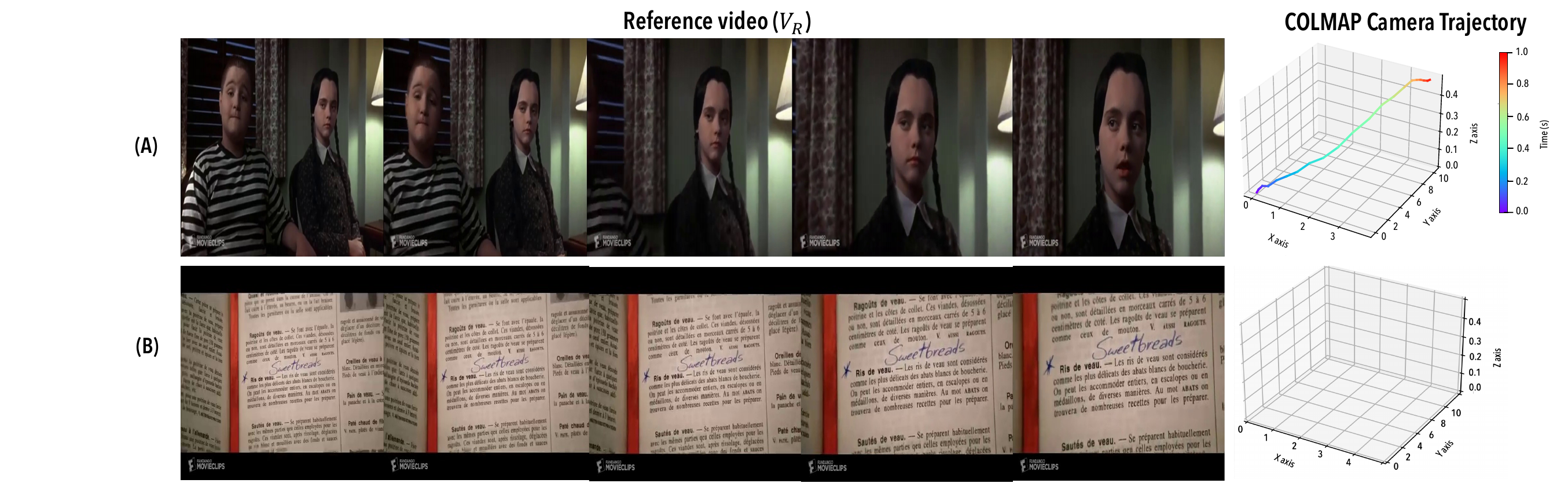}
    \caption{This showcases examples to depict scenarios where COLMAP can and cannot work reliably. In video (A), the camera effect observed is due to the explicit movement of the camera while in (B), the effect is obtained due to changes in camera focal length. COLMAP is not able to converge for videos like (B).}
    \label{fig:camera_traj_fail}
\end{figure*}
\begin{figure*}
    \centering
    \includegraphics[width=\textwidth]{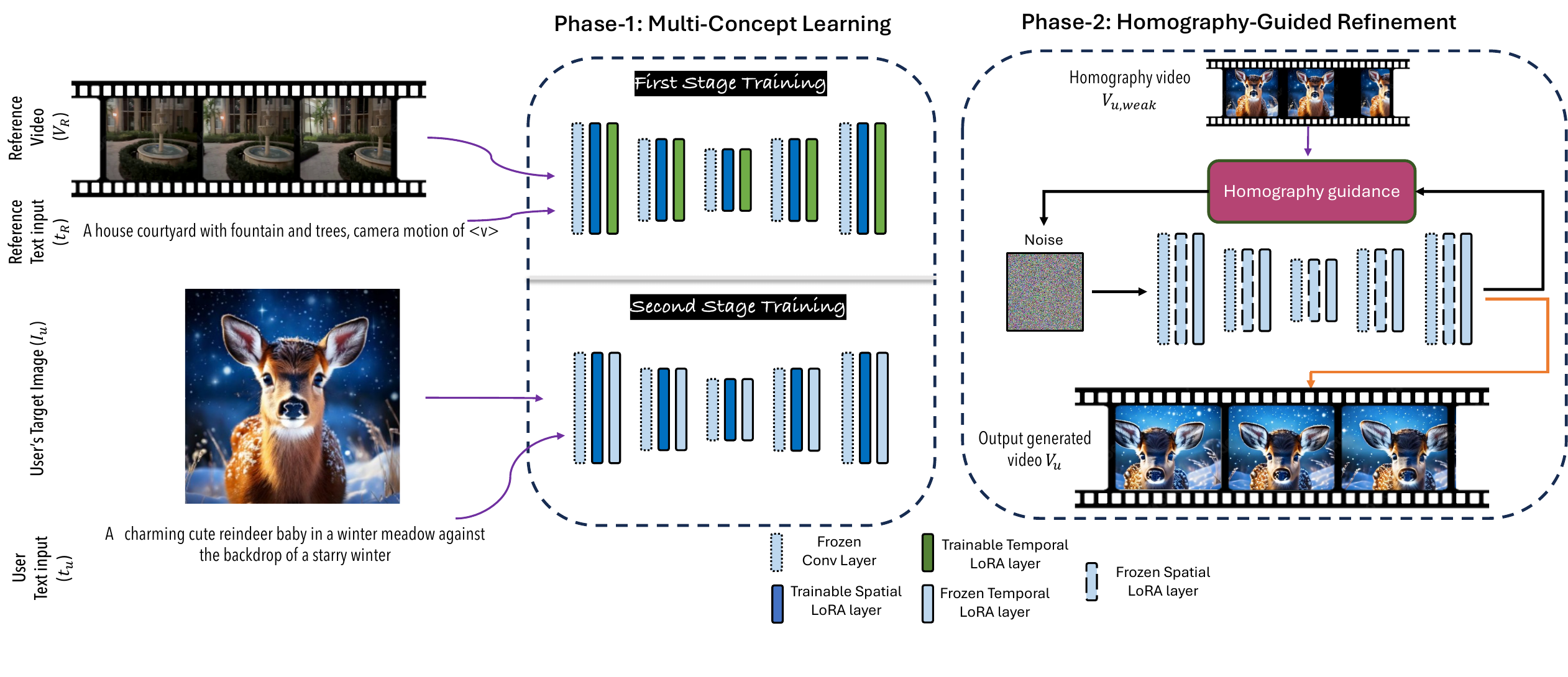}
    \caption{We present a zero-shot method to transfer camera motion visible in a reference video $V_R$ onto a user provided image $I_u$ to generate video $V_u$. It's a two-phase algorithm. The first phase involves learning multiple concepts associated with the spatial and temporal features of the reference video as well as the spatial characteristics of the user-provided image. We propose the use of a spatial-temporal orthogonality loss to better learn these concepts. The second phase consists of using a homography-based guidance to refine the generated video to preserve the scene in $I_u$ as well as the camera motion obtained from the reference video $V_R$.}
    \label{fig:overview_diagram}
\end{figure*}
\noindent\textbf{Problem Description.} Given a \textit{single} reference video $V_R$ (with prompt $t_R$), a user image $I_u$ (with prompt $t_u$), our goal is to generate a new video $V_u$ by transferring the camera motion from $V_R$ onto $I_u$ in a zero-shot manner. We use a pretrained text-to-video diffusion backbone, e.g., zeroscope~\cite{zeroscopev2}. $t_R$ is structured as a two-part, comma-separated prompt: visual content of $V_R$, followed by 'camera motion of $<v>$', where $<v>$ is a random token to handle arbitrary camera motions. $t_u$ describes the content of $I_u$.
\\

\noindent\textbf{Key Challenges.}
Approaching CMT by first extracting camera trajectories using methods like COLMAP and then integrating them with diffusion models~\cite{he2024cameractrl} has two key issues: \textit{(1)} Camera motions arise from explicit movement or focal length changes (e.g., dolly zoom). COLMAP \textit{struggles} with the latter (Fig.~\ref{fig:camera_traj_fail}). \textit{(2)} These methods \textit{require training specialized modules}, which is \textit{impractical for personalization} or limited-example scenarios. To overcome these, we propose a novel \textit{zero-shot} strategy.

\paragraph{{Method Overview.}} Our two-phase algorithm is shown in Fig.~\ref{fig:overview_diagram}: (i) \textit{Multi-concept finetuning} that learns to separate temporal and spatial features from $V_R$ and blend the spatial characteristics of $I_u$ using a novel orthogonality based regularizer, while allowing them to be integrated into a single controllable model; (ii) \textit{Homography-guided inference} to ensure $V_u$ preserves both the $I_u$ scene and the $V_R$ camera motion. We now discuss each of them in detail.

\subsection{Phase-1: Multi-Concept Finetuning}
\textbf{Network Structure:} Our zero-shot algorithm requires discerning three CMT-relevant concepts: $V_R$'s \textit{spatial characteristics}, $V_R$'s \textit{camera motion (temporal) features}, and $I_u$'s \textit{spatial characteristics}. We use a \textit{multi-concept} LoRA fine-tuning approach on the text-to-video backbone. LoRA~\cite{hu2021lora} enables efficient fine-tuning by constraining the weight update $W_0 + \Delta W = W_0 + BA$, where $B\in\mathcal{R}^{d\text{ x }r}$, $A\in\mathcal{R}^{r\text{ x }k}$, and rank $r$ is small:
\begin{equation}
 W_0 + \Delta W = W_0 + BA
\end{equation}
We use dual LoRA matrices: \textit{spatial LoRAs} in spatial self-attention layers to capture spatial characteristics, and \textit{temporal LoRAs} in temporal self-attention layers to model camera motion features. Cross-attention layers remain fixed.

\noindent\textbf{Training Strategy:} We employ a \textit{two-stage} training paradigm.

\noindent\textit{First stage training:} This differential training captures $V_R$'s spatial and temporal concepts. Temporal LoRAs are trained on all $V_R$ frames for comprehensive motion dynamics. Spatial LoRAs are trained on a single, random $V_R$ frame per step, promoting generalization. We use random masking for spatial LoRAs~\cite{zhuang2024copra} for regularization. The training loss is:
\begin{equation}
 \mathcal{L}_{first\_stage} = \mathcal{L}_{temporal} + \delta \mathcal{L}_{spatial}
\end{equation}
where $\mathcal{L}_{temporal}$ and $\mathcal{L}_{spatial}$ are the noise prediction losses~\cite{dhariwal2021diffusion} for their respective LoRAs, and $\delta \in (0, 1)$.

\noindent\textit{Second stage training:} This stage adapts the model to $I_u$ while maintaining $V_R$'s temporal characteristics. Spatial LoRAs are fine-tuned using $I_u$~\cite{asadi2024does}, leveraging the spatial-temporal relationship learned from $V_R$ in the first stage. To prevent spatial fine-tuning from suppressing temporal features, we introduce an orthogonality loss ($\mathcal{L}_{ortho}$). This constraint ensures spatial features remain orthogonal to temporal features, preserving the temporal dynamics while seamlessly integrating $I_u$'s spatial characteristics. The second stage training loss is:
\begin{equation}
 \mathcal{L}_{second\_stage} = \mathcal{L}_{spatial} + \lambda\mathcal{L}_{ortho}
\end{equation}
where $\lambda$ is a hyperparameter and:
\begin{equation}
 \mathcal{L}_{ortho} = \phi(W_{spatial\_LoRA}, W_{temporal\_LoRA})_k
\end{equation}
$\phi (x,y)_k$ is the inner product between the $k$ most significant values of $W_{spatial\_LoRA}$ and $W_{temporal\_LoRA}$ (the respective LoRA weight parameters).

\subsection{Phase-2: Homography Guided Inference}
We introduce a weak supervision strategy during the \textit{inference denoising} stage to synchronize motion and appearance in $V_u$. We leverage \textit{homography}~\cite{szeliski2022computer}, proven effective for novel view synthesis~\cite{kothandaraman2023hawki}, as a weak signal for how $I_u$'s scene elements \textit{might} change under the desired camera motion. Homography computes the transformation mapping one image onto another (rotation/translation) using key point matching (e.g., SIFT~\cite{lowe2004sift}) and RANSAC~\cite{derpanis2010overview}. 

Inference begins with the fine-tuned backbone using prompt $t_I = t_u + $`camera motion of $<v>$'. We compute frame-to-frame homography matrices $H_{i}$ (from $F_{R,i}$ to $F_{R,i+1}$) using the reference video $V_R$. We then generate a \textit{pseudo-weak} estimate, $V_{u,weak}$, starting with $I_u$ as the first frame ($F_{P,1}$). Subsequent frames are $F_{P,i+1} = H_i (F_P,i)$, where $i \in [1, N-1]$. At each diffusion denoising step (except the last), the predicted latents $z_{t}$ are adjusted~\cite{kothandaraman2023hawki, ho2022classifier} based on the latents $z_{P}$ corresponding to $F_{P}$:
\begin{equation}
\hat{z}_{t} = z_{t} - \lambda_{G} \nabla {(z_{t} - z_{P})}^2
\end{equation}

\subsection{CameraScore - A New Metric}
Popular motion transfer metrics like optical flow~\cite{beauchemin1995computation} and COLMAP~\cite{schoenberger2016sfm} are unsuitable for cross-scene CMT due to \textit{structural incompatibility, scaling differences, video quality dependency, and computation cost} (details in Appendix).

Therefore, we propose CameraScore, a homography-based metric for evaluating camera motion transfer across structurally different scenes:
\begin{equation}
 \frac{1}{N}\sum_{i=1}^{N-1} ||\mathcal{H}_{R_i} - \mathcal{H}_{G_i} ||^2
\end{equation}
where $\mathcal{H}_{R_i}$ is the homography matrix between the $i^{th}$ and $(i+1)^{th}$ frame of the reference video, and $\mathcal{H}_{G_i}$ is the same for the generated video.

\noindent\textbf{Why Homography?} Homography estimation effectively captures relative transformations (panning, tilting, zooming) between two planes, aligning perfectly with our goal to transfer holistic camera motion rather than pixel-level details (unlike optical flow). By considering homography between consecutive frame pairs, we capture local motions that approximate planar or rotational transformations. The cumulative effect represents the complex camera path, providing a valuable and computationally efficient metric compared to COLMAP or optical flow.

\section{Experiments and Results}
\label{sec:exp_results}

\subsection{Dataset}
\label{sec:dataset}
Since \textit{no} dataset exists specifically for zero-shot single-exemplar CMT, we curated a diverse dataset of 680 reference video and user scene combinations, comparable in size to works like DreamBooth~\cite{ruiz2023dreambooth} and MotionDirector~\cite{zhao2025motiondirector}. Reference videos, 2–4 seconds long, feature a wide range of cinematic motions: panning, zooming, tilting, dolly shots, and 3D rotations, capturing both \textit{subtle} and \textit{dynamic} styles from movies, documentaries, and animation. User scenes were generated using SOTA text-to-image models, spanning landscapes and urban settings. We used BLIP to caption images; three experienced annotators provided the content descriptions for $t_R$. This diversity ensures rigorous testing of our method's robustness and adaptability. We will release this dataset post-acceptance.

\subsection{Implementation Details}
\label{sec:implementation_details}
\noindent\textbf{Model Details.} We use \textit{zeroscope}~\cite{zeroscopev2} as the text-to-video backbone, generating 16 frames per sample.

\noindent\textbf{Training Details.} Models are fine-tuned on a \textit{single} A5000 GPU. Both training stages run for 150 epochs each, taking approximately 10 minutes total per sample to generate $V_u$. This two-stage adaptation is performed per reference video-user image pair and is intended as a personalized model creation step, similar to DreamBooth-style workflows, rather than a single universal model that instantaneously supports all motion and scenes. We use the Adam optimizer with a learning rate of $5e-4$. All code is implemented using Pytorch~\cite{imambi2021pytorch}.

\noindent\textbf{Metrics.} We evaluate using three metrics: DINOv2~\cite{oquab2023dinov2} (scene preservation/similarity to $I_u$), VideoCLIP~\cite{xu2021videoclip} (text consistency/alignment), and our proposed CameraScore (motion consistency with $V_R$).

\subsection{Comparison with Prior Work}
\label{sec:comparison}
Ours is the first paper to address \textit{zero-shot} CMT from a $\textit{single}$ video onto a target scene. SOTA foundation models do not natively support this task. We compare our approach against three related \textit{zero-shot} baseline categories: traditional homography, DreamBooth-style fine-tuning (using Tune-A-Video~\cite{wu2023tune} to introduce scene customization), and video diffusion-based object motion transfer (MotionDirector~\cite{zhao2025motiondirector}). Other recent camera-control methods based on explicit trajectories or NeRF-style inputs are not directly comparable in our single-exemplar, arbitrary-image setting. Therefore, we restrict our comparisons to only these zero-shot diffusion-based baselines that can be instantiated under the same assumptions.

\begin{figure}[b]
\includegraphics[width=\linewidth]{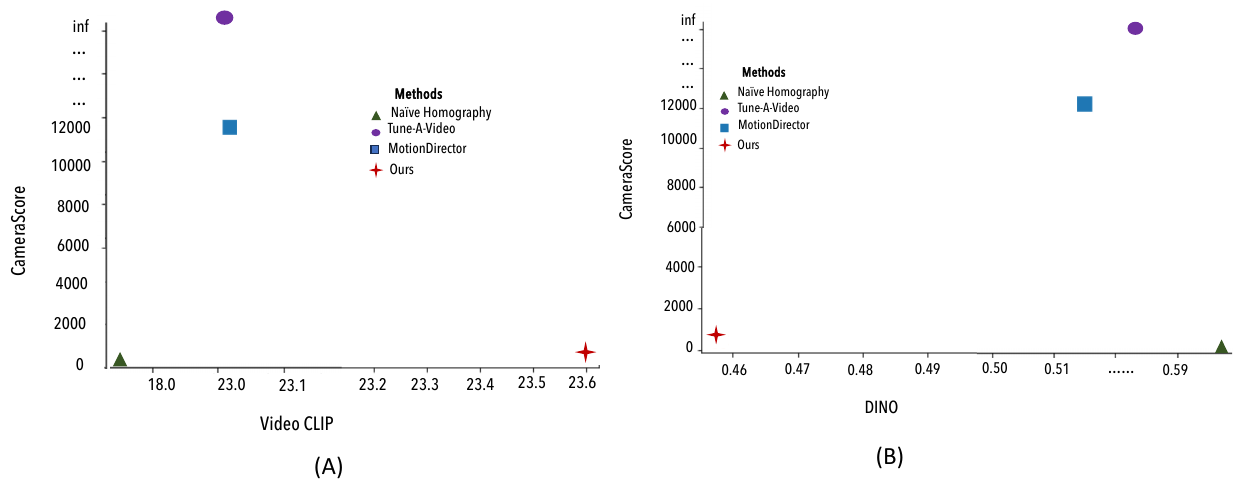}
\caption{Graph (a) shows the results of the plot of Video CLIP scores with CameraScore for the different diffusion methods being compared. Graph (b) shows the plot of DINO scores with CameraScore for the different diffusion methods being compared. Our approach achieves the \textit{best trade-off}.}
 \label{fig:exp_results}
 \end{figure}

Fig.~\ref{fig:exp_results} presents the \textit{quantitative comparisons}. The homography baseline accurately captures motion but suffers from outpainting artifacts. Tune-A-Video achieves high scene consistency but minimal camera movement. MotionDirector introduces slight motion but struggles with integrating $I_u$ spatial characteristics with $V_R$ temporal features, often suppressing temporal dynamics.

\begin{figure*}[t]
 \centering
 \includegraphics[width=\textwidth]{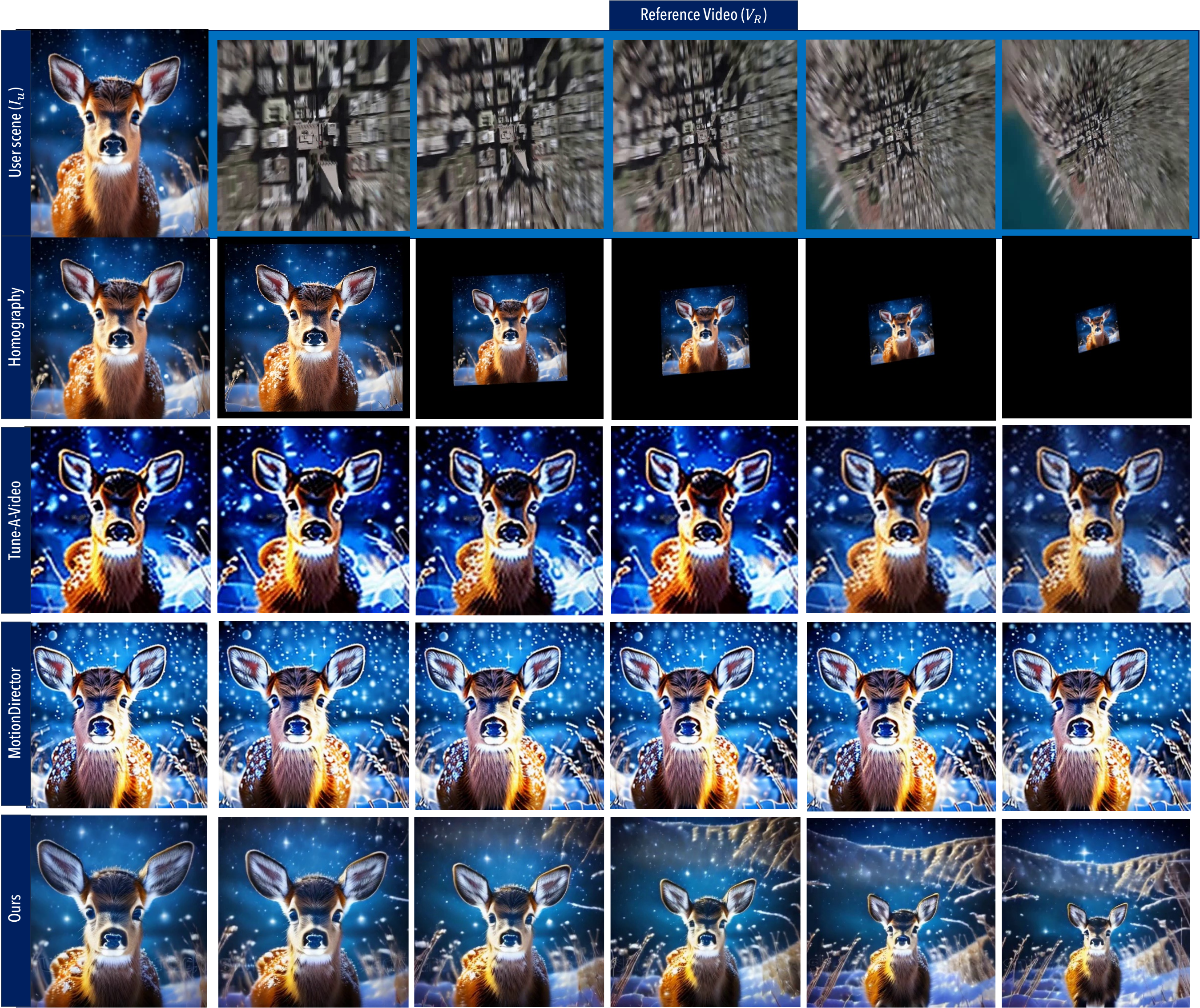}
 \caption{Qualitative results of our method, demonstrating clear improvements over prior work in transferring camera motion, while preserving scene content (\href{https://www.youtube.com/watch?v=NLeTEjx3njg}{Reference video}). \textit{More results in the appendix.}}
 \label{fig:qualtiative_results1}
\end{figure*}

In contrast, Our method achieves the best trade-off. We achieve a higher VideoCLIP score than MotionDirector and Tune-A-Video, indicating superior textual consistency and better adherence to the motion prompt. Our lower CameraScore compared to MotionDirector signifies better motion transfer fidelity. We also attain a higher DINO score than MotionDirector, showcasing improved scene preservation with respect to $I_u$. Tune-A-Video, which lacks motion, has an inflated DINO score as it simply replicates $I_u$. Qualitative results (Fig.~\ref{fig:qualtiative_results1}) further validate our approach's effectiveness in transferring camera motion while preserving scene integrity.

\subsection{Ablation Experiments}
\label{sec:ablation}
We ablate the impact of three key components: USC (\textit{U}ser \textit{S}cene \textit{C}haracteristics learning: $I_u$ spatial feature learning in Phase-1, excluding orthogonality loss), ORTHO (orthogonality loss function to prevent $I_u$ spatial features from suppressing $V_R$ temporal features), and HG (\textit{H}omography \textit{G}uidance during Phase-2 inference).

\begin{figure}[b]
 \includegraphics[width=\linewidth]{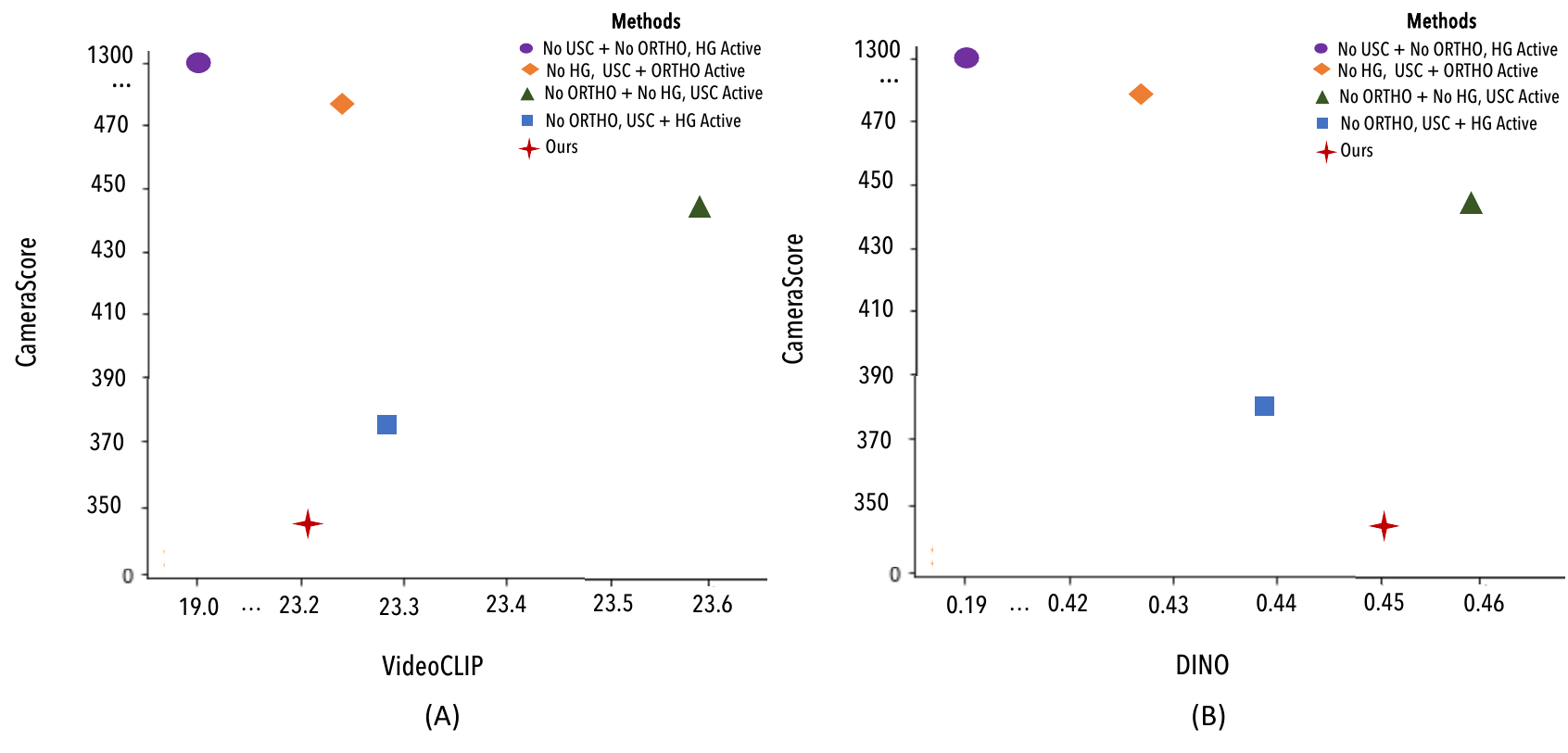}
 \caption{Graph (A) illustrates the relationship between CameraScore and DINO, while Graph (B) shows the relationship between CameraScore and VideoCLIP. These plots are derived from the ablation experiments conducted to evaluate the significance of various key components in our proposed approach.}
 \label{fig:ablation_studies}
 \end{figure}

Fig.~\ref{fig:ablation_studies} shows the results. Our full approach achieves the \textit{best} trade-off. Whe USC is removed (removing $I_u$ information), the network fails to transfer motion effectively, resulting in a significantly higher CameraScore; HG becomes ineffective without the user scene as a base. When only USC is used, CameraScore deteriorates compared to when ORTHO and HG are included, confirming that without ORTHO, temporal features are disrupted despite better spatial feature capture. Without ORTHO, the camera motion transfer is disrupted as expected. Overall, our findings demonstrate that USC, ORTHO, and HG are complementary and essential for effective motion transfer aligned with the reference video while preserving spatial scene integrity.


\begin{figure}
\centering
\includegraphics[width=\columnwidth]{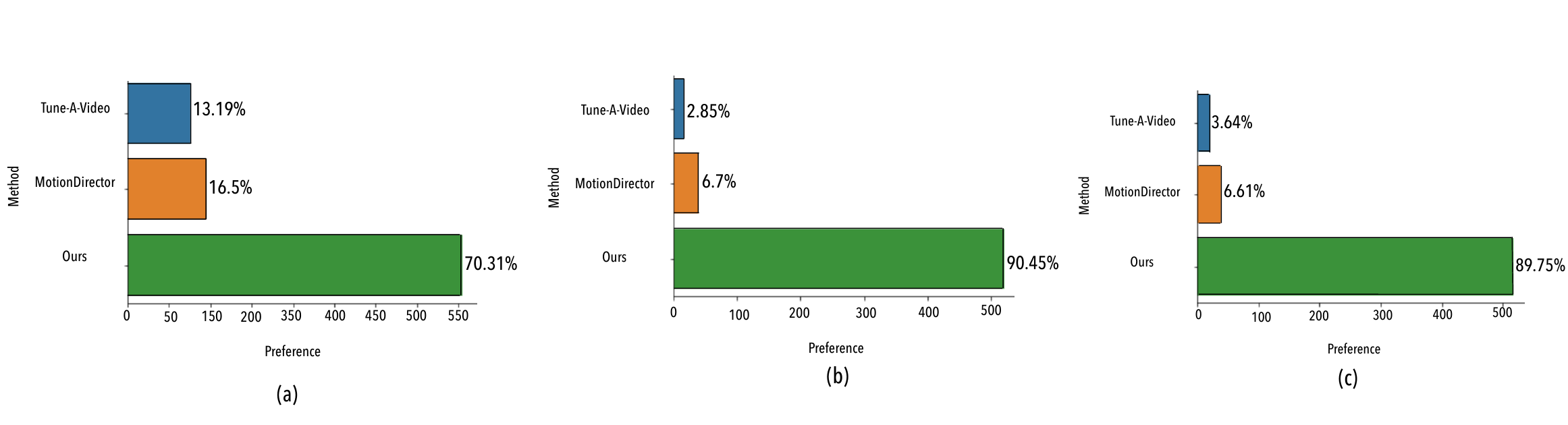}
\caption{Graph (a) shows participants evaluating the similarity between the generated video scene and the user-provided scene. Graph (b) depicts participants' preferences for the output video with the desired camera motion among the three methods. Graph (c) indicates participants' choice of the overall best output among the different methods.}
\label{fig:User-Study}
\end{figure}

\subsection{User Study}
To evaluate the effectiveness and usability of our method, we conducted two complementary user studies. The first focuses on video quality, assessing how accurately our system reproduces camera motions and preserves user-provided image characteristics. The second focuses on interface usability, comparing different interaction paradigms for directing camera motion. Together these studies provide a holistic understanding of both algorithmic performance and human-centered creative support. 
\subsubsection{Video Quality Study}
The goal of this study was to validate the fidelity of camera motion transfer while maintaining the visual integrity of the user's scene. 
\paragraph{Participants}
We recruited $72$ participants via an online crowdsourcing platform to evaluate the algorithmic quality of our system's output. The participant pool was diverse in terms of gender and age to reflect a broad, general user base representative of typical video consumers. Specifically, 50\% identified as male, 47\% identified as female, and 3\% preferred not to disclose or identified outside the gender binary. The age group spanned 18 to 65+, with most participants between 24 and 44 years old. 

None of the participants were required to have prior expertise in video editing or cinematography, supporting the system's accessibility goals. While not professionally trained, the participants reported frequent exposure to diverse video content in daily life, equipping them with the informal experience necessary to provide meaningful perceptual judgments relevant to non-expert users. 
\paragraph{Procedure} To assess the algorithmic quality of our algorithm's camera motion transfer, participants completed a forced-choice comparative evaluation. Each participant was made to evaluate 8 sets of test cases covering a diverse range of camera motions. 
Each set included: 
\begin{itemize}
    \item A reference video demonstrating the desired camera motion
    \item The target user image, and 
    \item Outputs from three methods: MotionDirector~\cite{zhao2025motiondirector}, Tune-A-Video (DreamBooth-modified)~\cite{wu2023tune}~and our method. 
\end{itemize}
Participants were asked to compare and choose their preferred output corresponding to each of the following three criteria:
\begin{enumerate}
    \item \textbf{Scene similarity} - how well the generated video preserves the user's image characteristics
    \item \textbf{Camera motion fidelity} - how closely the camera movement matches the reference video, and 
    \item \textbf{Overall performance} - which output they considered best overall.
\end{enumerate}
A brief tutorial with examples familiarized participants with camera motion transfer concepts and evaluation criteria. To ensure unbalanced judgments, we randomized the order of output options within each test set to prevent positional bias. Additionally, the presentation order of the test sets themselves was randomized for each participant.
\paragraph{Statistical Analysis}
For each participant and criterion, we determined which method they selected the most frequently across the 8 trials. These participant-level preferred methods yielded the counts reported in Table~\ref{tab:algorithm_stats} and served as the inputs to our inferential tests. 
Participant preferences were analyzed using a chi-square goodness-of-fit test to evaluate whether the distribution of choices differed significantly from chance among the three competing methods: MotionDirector, Tune-A-Video(DreamBooth modified) and our method. Under the null hypothesis, each method was equally likely to be preferred at the participant level, with a probability of $1/3$ for each method. When the chi-square test indicated a significant deviation from equal preference, we conducted pairwise post-hoc comparisons via binomial tests between our method and each baseline method to identify which pairs contributed to the overall preference differences. Bonferroni correction was applied to adjust for multiple comparisons, setting a significance threshold of $\alpha=0.025$ for the two post-hoc tests per criterion. Effect sizes were calculated using \text{Cram\'{e}r's} V to quantify the strength of the association between method and preference, with established ndicating small (0.1), medium (0.3) and large (0.5) effects. All statistical analysis were conducted using Python 3.11 and SciPy 1.10.
\paragraph{Results}
Table~\ref{tab:algorithm_stats} summarizes participant preferences across three criteria: scene similarity, camera motion fidelity and  overall performance. Our method was the predominant choice for all criteria, selected by $70.31\%, 90.45\%$ and $89.75\%$ of participants respectively (Figure~\ref{fig:User-Study}).
\begin{table*}[ht]
    \centering
    \caption{Statistical Analysis of Video Quality User Study}
    \resizebox{\textwidth}{!}{%
   \begin{tabular}{cccccccc}
    \toprule
    \textbf{Criterion} & \textbf{Tune-A-Video} & \textbf{MotionDirector} & \textbf{Our Method} & \textbf{Chi-square ($\chi^2$)} & \textbf{p-value} & \textbf{\text{Cram\'{e}r's}} & \textbf{Interpretation}\\
    \midrule
    Scene Similarity & $9 (13.19\%)$ & $12 (16.5\%)$ & $51 (70.31\%)$ & $45.75$ & $< 0.001$ & $0.56$ & Large effect\\
    Camera Motion Fidelity & $2 (2.85\%)$ & $5 (6.7\%)$ & $65 (90.45\%)$ & $105.25$ & $<0.001$ & $0.85$ & Large effect\\
    Overall Performance & $3 (3.64\%)$ & $5 (6.61\%)$ & $64 (89.75\%)$ & $100.08$ & $<0.001$ & $0.83$ & Large effect\\
    \bottomrule
    \end{tabular}
    }
    \label{tab:algorithm_stats}
\end{table*}
Post-hoc binomial tests with Bonferroni correction confirmed that our method was significantly preferred over both Tune-A-Video and MotionDirector across all criteria (all adjusted $p<0.01$).
We also observe that the method with the lowest CameraScore (ours) is the one most preferred for motion fidelity in the user study (90.45\%), providing qualitative evidence that CameraScore is directionally consistent with human judgments for our scenarios.

\subsubsection{User Interaction Study}
To understand how effectively our system supports creative workflows, we conducted a comparative study of three interface paradigms for replicating camera motions: (1) text prompt-based controls, (2) preset-based motion panels, and (3) reference video-driven interface (ours). The study aimed to evaluate usability, efficiency, and participants' ability to achieve their intended cinematic outcomes. By observing how participants interact with each interface and collecting subjective feedback, we assessed which design best enables creators to explore, iterate, and express their creative vision.

\paragraph{Participants}
We recruited 12 participants (7 female, 5 male) with ages ranging from 19 to 52 years (median = 29.5, SD = 7.42). Participants reported their experience with video editing tools on a 7-point scale, with a median experience level of 4 (SD = 1.26). Their experience with video generation tools was lower, with a median of 2 (SD = 1.35). Participants were recruited through university mailing lists and social media, and received compensation for their time.

\paragraph{Apparatus and Materials}
Participants used three different video generation interfaces:
\begin{itemize}
\item \textbf{Text}: Frame-to-video generation using text prompts with Flow (Veo3~\cite{google_veo3_2025})
\item \textbf{Preset+Text}: Frame-to-video with text prompts plus one of 13 predefined camera motion presets using Veo2~\cite{veo2}
\item \textbf{Reference Video (Ours)}: Our developed interface using reference video-driven motion transfer
\end{itemize}
We selected Google's Flow~\cite{google_flow_2025} as our baseline text-based interface due to its state-of-the-art performance in video generation and its widespread adoption in the creative community. Flow represents the current industry standard for text-to-video generation, making it an appropriate comparison point for evaluating alternative interface paradigms. The preset-based condition using Veo2 was included to assess whether structured motion controls could bridge the gap between pure text descriptions and reference-driven approaches.

Each session used standardized input images and inspiration videos to simulate scenarios where users have a specific video outcome in mind.

\paragraph{Procedure}
The study followed a within-subjects design with counterbalanced presentation order. Each session began with introductions and consent procedures, followed by an explanation of the study goals and tasks. Participants were instructed to focus on recreating camera movements and effects rather than altering video content, and to disregard video resolution or quality in favor of achieving desired motion effects.

For each interface condition, participants were presented with a creative scenario simulating a common workflow where content creators seek to replicate cinematic effects they have observed. Specifically, participants were shown an input image as their starting frame and an inspiration video demonstrating a desired camera motion or cinematic effect. This setup reflects real-world situations where creators encounter compelling visual content and wish to incorporate similar stylistic elements into their own work.

Participants completed three distinct camera motion tasks in counterbalanced order: (1) an orbit motion involving 3D rotation around a central subject, (2) a zooming motion with varying speed dynamics, and (3) a 2D rotation motion. These tasks were selected to represent common cinematic techniques that vary in complexity and require different levels of spatial understanding and motion control. Participants were then asked to achieve the demonstrated camera motion using the available interface, with unlimited iterations for the Reference Video condition and a maximum of 5 iterations for Text and Preset+Text conditions due to platform token limitations. Participants could iterate until satisfied with their results.

Throughout each trial, we recorded participant's screen to measured task completion time and iteration count from initial idea formulation to final satisfactory outcome. After completing each condition, participants filled out questionnaires including the NASA Task Load Index (TLX)~\cite{hart1988development}, System Usability Scale (SUS)~\cite{brooke1996sus}, and 7-point Likert scales assessing satisfaction with their generated video and preference for the given method.

\paragraph{Measures}
We collected both objective and subjective measures to comprehensively evaluate the effectiveness of each interface paradigm. Our objective measures included the number of iterations required to achieve satisfactory results, which served as an indicator of interface efficiency and ease of achieving desired outcomes. We also measured task completion time from initial interaction to final satisfactory outcome, providing insights into the cognitive load and learning curve associated with each interface. Screen recordings of user interactions were captured to enable detailed analysis of user behavior patterns, interaction strategies, and potential usability issues.
For subjective assessment, we employed the NASA Task Load Index (TLX) to evaluate perceived workload across six dimensions: mental demand, physical demand, temporal demand, performance, effort, and frustration. This multidimensional assessment helps understand the cognitive burden imposed by each interface design. We administered the System Usability Scale (SUS) to obtain standardized usability scores, enabling comparison with established usability benchmarks and between our three interface conditions. Additionally, we used 7-point Likert scales to assess participants' satisfaction with their generated video outcomes and their overall preference for each method, capturing the subjective quality of the creative experience. Post-session interviews provided qualitative insights into participants' reasoning, preferences, and suggestions for improvement, offering deeper understanding of the quantitative findings.
 
\paragraph{Results}

\begin{figure}
    \centering
    \includegraphics[width=\columnwidth]{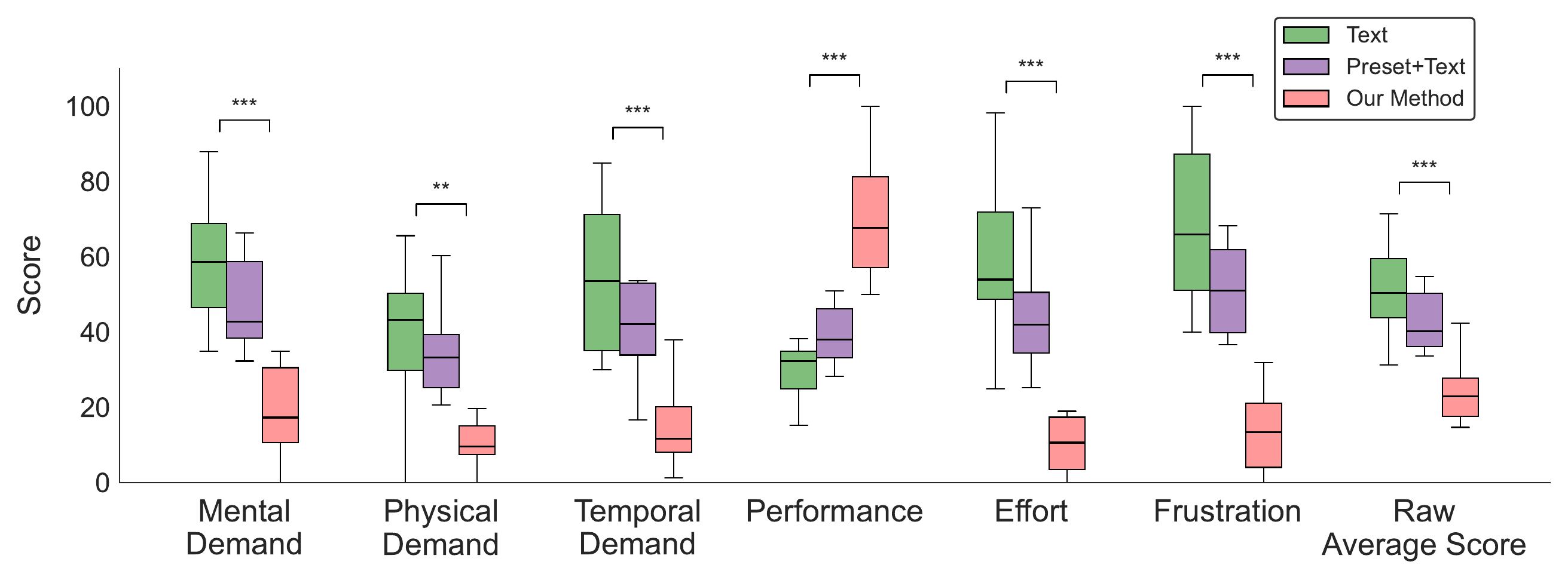}
    \caption{NASA TLX Results}
    \label{fig:result-nasa}
\end{figure}

\begin{figure}
    \centering
    \includegraphics[width=1.1\columnwidth]{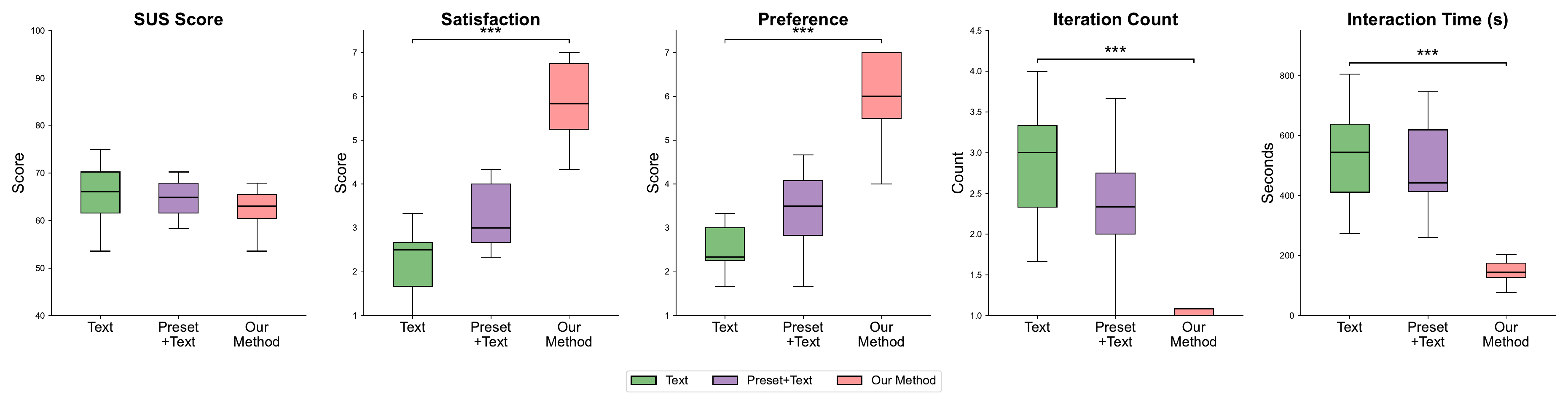}
    \caption{Interaction Time}
    \label{fig:result-rest}
\end{figure}

\paragraph{NASA Task Load Index (TLX)}
The NASA-TLX assessment revealed significant differences across all six dimensions of cognitive workload between the three methods. Our Method consistently demonstrated the lowest workload scores, indicating superior usability compared to both Text and Preset+Text approaches. (See Figure~\ref{fig:result-nasa})

Mental demand showed significant differences between methods (ANOVA: $F = 26.62$, $p < 0.001$), with Our Method requiring substantially less mental effort than the baseline approaches. Physical demand also differed significantly (Friedman test: $\chi^2 = 11.64$, $p = 0.003$), though all methods scored relatively low on this dimension as expected for text-based interfaces.

Temporal demand exhibited strong significant differences (Friedman test: $\chi^2 = 18.57$, $p < 0.001$), with Our Method allowing users to complete tasks with less time pressure. Performance ratings differed significantly between methods (Friedman test: $\chi^2 = 21.57$, $p < 0.001$), with Our Method receiving the highest performance scores, indicating users felt they accomplished their goals more effectively.

Effort requirements showed significant variation (Friedman test: $\chi^2 = 20.67$, $p < 0.001$), with Our Method requiring notably less user effort. Frustration levels differed markedly between approaches (ANOVA: $F = 42.13$, $p < 0.001$), with Our Method producing the lowest frustration scores. The overall NASA-TLX scores (Raw Average Score) demonstrated significant differences (ANOVA: $F = 23.60$, $p < 0.001$), confirming Our Method's superior workload profile across all dimensions. 

\paragraph{System Usability Scale (SUS)}
SUS scores did not reveal significant differences between the three methods (ANOVA: $F$ not reported, $p = 0.230$). See Figure~\ref{fig:result-rest}). All methods achieved moderate usability scores (Text: $M = 66.77$, $SD = 8.97$; Preset+Text: $M = 64.58$, $SD = 3.90$; Our Method: $M = 62.40$, $SD = 3.98$), suggesting that while users found all approaches reasonably usable, the SUS instrument may not have been sensitive enough to detect the performance differences observed in other measures. 

\paragraph{User Satisfaction and Preference}
User satisfaction ratings differed significantly between methods (ANOVA: $p < 0.001$), with Our Method receiving substantially higher satisfaction scores ($M = 5.92$, $SD = 0.95$) compared to Preset+Text ($M = 3.25$, $SD = 0.77$) and Text ($M = 2.31$, $SD = 0.72$). This pattern indicates that users found Our Method considerably more satisfying to use.

User preference showed a similar pattern (ANOVA: $p < 0.001$), with Our Method strongly preferred ($M = 5.97$, $SD = 1.06$) over both Preset+Text ($M = 3.36$, $SD = 0.93$) and Text ($M = 2.42$, $SD = 0.65$). These results suggest that users not only performed better with Our Method but also preferred the interaction experience it provided.

\paragraph{Task Efficiency Metrics}
Iteration count analysis revealed significant differences between methods (Friedman test: $p < 0.001$). Our Method required significantly fewer iterations to complete tasks ($M = 1.11$, $SD = 0.22$) compared to both Preset+Text ($M = 2.36$, $SD = 0.73$) and Text ($M = 2.94$, $SD = 0.71$). This reduction in required iterations suggests Our Method enabled users to achieve their goals more directly with less trial-and-error.

Interaction time showed dramatic differences between approaches (ANOVA: $p < 0.001$). Our Method required substantially less time to complete tasks ($M = 159.17$ seconds, $SD = 56.43$) compared to Preset+Text ($M = 490.47$ seconds, $SD = 142.51$) and Text ($M = 531.00$ seconds, $SD = 169.60$).

The results demonstrate that Our Method consistently outperformed both baseline approaches across multiple dimensions. Users experienced lower cognitive workload, higher satisfaction and preference, and achieved tasks more efficiently with fewer iterations and significantly less time. While SUS scores showed no significant differences, the convergent evidence from NASA-TLX, satisfaction measures, and objective performance metrics strongly supports the effectiveness of Our Method for this task domain.

These user studies serve as a \textit{complementary evaluation channel} to CameraScore especially for scenes where planar homography assumptions are violated or where subtle cinematic qualities are hard to capture with geometric metrics alone. 

\section{Discussion}
In this section, we reflect on our findings from the design and development of our system to seamlessly transfer camera motions from a reference video to any scene of the user's choice. 
\paragraph{\textbf{Automating Complex Technical Tasks to Foster Accessibility.}} Our findings demonstrate that embedding deep automation for technically challenging tasks such as accurate camera motion estimation substantially lowers cognitive barriers for users. By abstracting away complicated mathematical and calibration procedures behind intelligent AI models, systems like ours can empower novices and creators of varying expertise to produce high-quality, professional-grade content effortlessly. This highlights a critical design principle:\textit{effective AI-enhanced tools must prioritize invisibility and simplicity in the user interface while delivering sophisticated capabilities under the hood.} Future designs should ensure that powerful automation complements rather than complicates the creative process, thereby reducing friction and inviting broader adoption. 
\paragraph{\textbf{Harnessing Reference-Driven and Incremental Interaction Paradigm.}} In our interaction study, user preferences strongly favored exemplar-based workflows. This emphasizes the need for providing \textit{concrete visual or textual anchor points} facilitates more intuitive and controllable content generation. Users benefit from workflows that allow them to iteratively adjust AI-generated content through small, understandable steps, rather than relying on opaque, fully automated outputs. This incremental, transparent process makes the creative journey more accessible, and predictable. This enables users to maintain control, build confidence and continuously guide AI towards their desired outcome. Designers should move beyond fully automated \textit{black-box} solutions toward collaborative interaction models where users play an active role in guiding the generative behaviors. Such hybrid approaches balance user agency with AI efficiency, enhancing trust and satisfaction.
\paragraph{\textbf{Building Modular, Extensible, and Future Proof Architecture.}} 
Our system's successful integration of reference video based camera motion inference with generative AI pipelines highlights the importance of \textit{modular and extensible system architectures}. Such designs empower developers to fluidly incorporate state-of-the-art algorithms, respond to emerging user requirements, and maintain system evolvability without disrupting user experience continuity. Prioriziting architecture strategies that enable incremental innovation and cross component synergy will be crucial for sustaining the long-term relevance and usability of AI-powered creative tools. 
\section{Limitation}
Our system represents a significant step forward in zero-shot reference video based camera motion transfer. However, several limitations remain. Technically, our reliance on homography-based guidance can be susceptible to inaccuracies in the presence of substantial foreground object motions within reference videos. This limitation restricts motion transfer fidelity when dynamic objects disrupt stable camera motion estimation. Additionally, our system currently supports camera motion transfer only onto static images rather than full video sequences, limiting its applicability to scenes with moving subjects or complex temporal dynamics. Furthermore, the quality and diversity of generated videos output are inherently bounded by the capabilities of the underlying pretrained generative model, constraining the complexity of camera motions and scene content that can be realistically synthesized. Although we implemented our method on Zeroscope due to availability, the dual-LoRA and homography-guidance strategy is model-agnostic and could be applied to other text-to-video backbones; we leave systematic evaluation across multiple models for future work. 

From an interface perspective, although our design lowers barriers by enabling intuitive camera motion specification via single reference videos, it lacks flexibility for more nuanced creative control. Users cannot blend multiple reference camera motions, adjust the point of application of camera motion (for example, allowing users to decide about which object in the scene, should a rotational camera motion occur) or finetune camera trajectories interactively. Additionally our user study focused on a small group of mostly non-expert creators. A more diverse participant base including professional filmmakers and active content creators may reveal additional usability needs and interaction challenges and we plan to evaluate the interface with expert users and more diverse content types in future work. 

\section{Future Work}
Building on these limitations, future research can pursue several promising directions. Enhancing camera motion transfer robustness by integrating object-aware scene understanding, potentially through large language models (LLM) enhanced video analysis could improve handling dynamic objects in reference videos. Extending our system to support full user videos rather than static images would enable richer, temporally coherent personalization workflows. Advances in generative video foundation models open opportunities for our camera motion transfer algorithm to evolve as a modular, plug and play component that leverages improved motion complexity and video fidelity. 

From an interaction perspective, users indicated that supporting fusion of multiple reference videos could enrich creative control further. While this capability to the best of our knowledge is not yet algorithmically implemented, designing interfaces that allow users to blend, select or weigh multiple references videos to be incorporated in the same video represents a valuable direction for future interface and algorithm co-design. Additional enhancements could include interactive editing controls to adjust the point of camera motion application and trajectory fine-tuning, as well as recommendation systems for reference video selection. Conducting broader user studies with diverse creator population will further inform the design of expressive accessible tools that align well with varied workflows and expertise levels. 

\section{Conclusion}
\label{sec:conclusion}

We designed and developed a system for reference based camera motion transfer that enables creators to animate static images with cinematic camera movements drawn from a \textit{single} reference video. The system leverages pretrained text-to-video diffusion models and incorporates an inference-time optimization strategy using dual LoRA networks, and homography guidance to achieve coherent and personalized motion effects. Our evaluations included both algorithmic studies, introducing a new homography-based metric and large scale user comparisons and interaction studies comparing interface paradigms for directing camera motion. Results show that our system not only reliably transfers the desired camera motion and preserves user's desired scene compared to prior methods but also enhances usability and creative control for non-expert creators. We hope this work catalyzes future research on human-centered generative video systems that expand the expressive possibility of storytelling.
\clearpage
\begin{appendices}
\section{Prior Works}
We provide a feature-by-feature comparison of prior works in this domain in Table~\ref{tab:prior_works_comparison}. We use~\checkmark~to mark the presence of a feature and~\xmark~to mark the absence.
\\
\begin{table*}
\caption{Comparison of our work with related prior works}
\label{tab:prior_works_comparison}
\resizebox{\linewidth}{!}{
\begin{tabular}{cccccc}
\hline \\
\textbf{Method} & \textbf{Zero-Shot strategy} & \textbf{Single Reference Video (input)} & \textbf{User-specified Image (input)} & \textbf{3D data (input)} & \textbf{Output Feature}s \\
\\\hline\\ 
Video Foundation Models~\cite{moviegen_meta, blattmann2023stable, wang2023modelscope} & \xmark & \xmark & \xmark & \xmark & Lacks the ability to transfer camera motion from a video onto another\\
\\\hline\\
AnimateDiff-based works~\cite{guo2023animatediff, hu2024motionmaster} & \xmark & \xmark & \xmark & \xmark & Text to video models that require training a control net and therefore, extensive data corresponding to it.\\ 
\\\hline\\
Uses 3D camera trajectories/3D models~\cite{he2024cameractrl,zhang2024recapture} & \xmark & \xmark & \textbf{\checkmark} & \checkmark & Requires extensive data for training a separate motion and camera control modules or 3D structure-aware modules. Additionally requires extraction of camera trajectories or generation of 3D models\\
\\\hline\\
Homography & \checkmark & \checkmark & \checkmark & \xmark & Can't mimic non-planar camera motions effectively. Black spaces in regions that are outside the boundaries of the user-specified image.\\
\\\hline\\
Tune-A-Video based works~\cite{wu2023tune, ren2024customize} & \checkmark & \checkmark & \xmark & \xmark & Text to image model extended for video generation. Targeted towards object motions. Output video content based on the text prompt provided.\\
\\\hline\\
DiTFlow based works\footnote{Pondaven, Alexander, et al. "Video motion transfer with diffusion transformers." Proceedings of the Computer Vision and Pattern Recognition Conference. 2025.} & \checkmark & \checkmark & \xmark & \xmark & Can't transfer to images. Video generated using only text prompts.\\
\\\hline\\
MotionDirector~\cite{zhao2025motiondirector} & \checkmark & \checkmark & \xmark\textsuperscript{*} & \xmark & Focuses on object motions. Primarily, output video content based on the text prompt provided \\
\\\hline\\
\textbf{Ours} & \checkmark & \checkmark & \checkmark & \xmark & Require only the reference video and user-specified image with the target being transfer of camera motions\\

\\\hline 
\end{tabular}
}
\end{table*}
For experiments, as discussed in Sec. 4.3, we compare our method only with prior zero-shot-based works like Homography, Tune-A-Video and MotionDirector. We detail these methods and highlight any modifications made to ensure a fair and consistent evaluation below.

\noindent\textbf{Naive Homography.} A fundamental approach to understanding scene changes due to camera motion is based on homography. This involves computing homography matrices from transitions observed in the reference video \( V_R \) and applying the same sequence to the target image \( I_u \). Specifically, we define:  

\[
V_u(t) = \mathcal{H}(V_R, t)I_u(t)
\]

where \( \mathcal{H}(V_R, t) \) represents the homography matrix computed at time step \( t \) between consecutive frames of \( V_R \), which is then applied to \( I_u(t) \), the corresponding state of \( I_u \).  
Given the central role of homography in our method, we also investigate the impact of integrating diffusion models for this task. This experiment evaluates the effect of excluding diffusion models on the generated outputs.\\

\noindent\textbf{Tune-A-Video (Dreambooth-modified).} In our method, we have used a video diffusion model. However, these are part of very recent developments within the diffusion literature. Text-to-image models have improved multi-fold since their inception. Tune-A-Video showcased extending these models to generate videos based on the text prompt provided. This method cannot be directly applied to our task, as we need the motion to be applied to a specific scene. We, therefore, modify the method to incorporate a Dreambooth-style~\cite{ruiz2023dreambooth}~training mechanism. This involves finetuning the Tune-A-Video 3D UNet with the video and its corresponding text, followed by finetuning it for the user image $I_u$. We have used the publicly available implementations of this work and DreamBooth to implement and test this case.\\ 

\noindent\textbf{MotionDirector~\cite{zhao2025motiondirector}.} In contrast to Tune-A-Video, this work does the same task of transferring object motions observed in a reference video but using a video diffusion model. For scene customization, they suggest finetuning two separate UNets - one on the reference video and one on the desired scene. Following this, they replace the spatial LoRA layers in the UNet finetuned on the reference video with the spatial LoRAs of the UNet finetuned on the image. We use the publicly available implementation for this work.

\section{CameraScore}
In Section 3.3, we introduced \textit{CameraScore}, a novel metric designed to evaluate the quality of camera motion generated by a method in comparison to the motion in the reference video. A \textit{lower score} indicates better performance. 
We deliberately avoid assessing camera motion using the estimated 3D trajectory typically obtained via COLMAP, as it proves unreliable or infeasible under specific conditions. For example, when the camera achieves a zoom effect solely by adjusting the focal length without physical movement, or when the scene structure is predominantly flat (i.e., coplanar), COLMAP struggles to generate accurate trajectories.

\begin{figure*}
    \centering
    \includegraphics[width=1\linewidth]{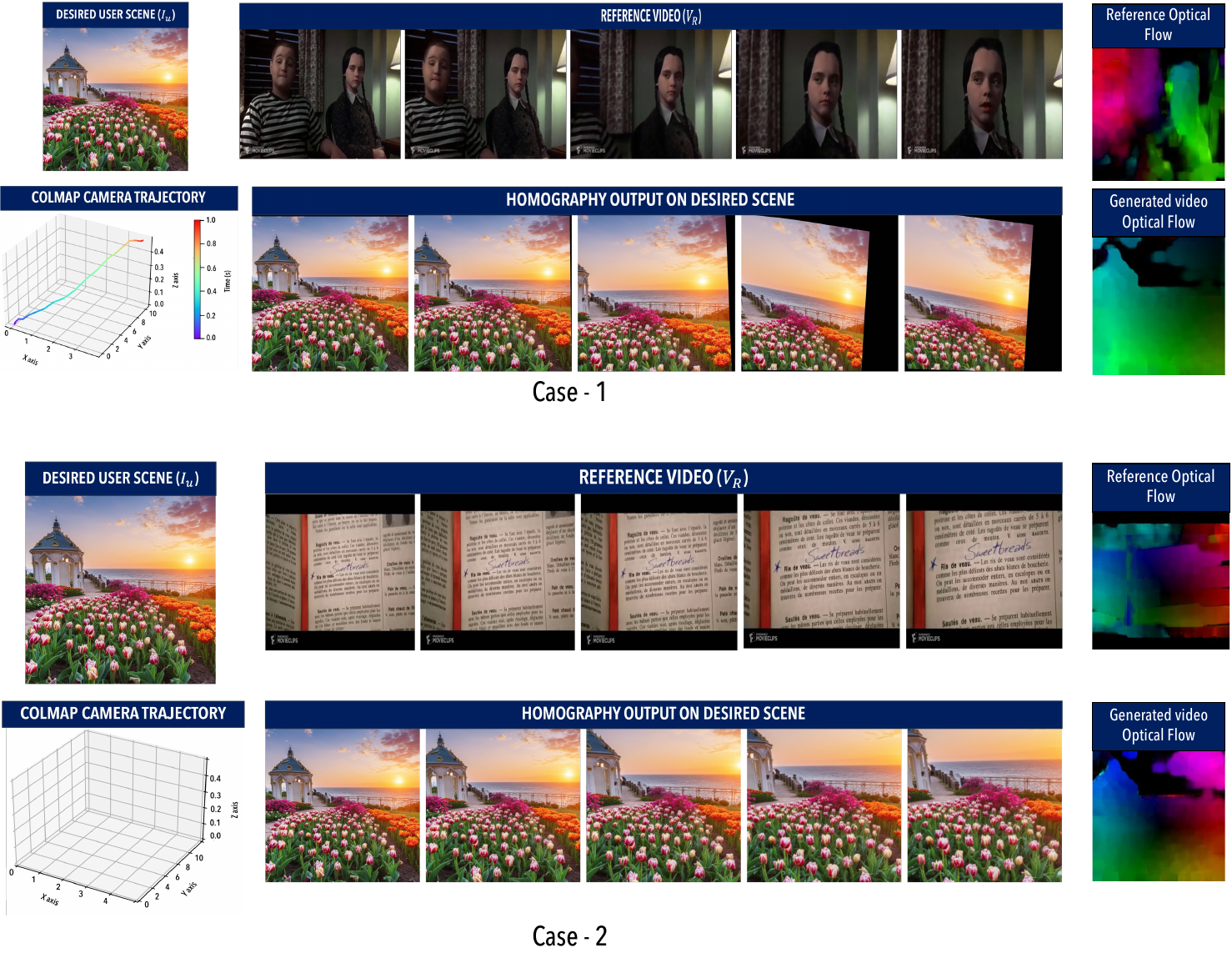}
    \caption{COLMAP and optical flow are not reliable metrics for comparing the camera motions observed in the reference and the generated videos. COLMAP fails to generate the camera trajectory in Case-2. Comparing the optical flows would be challenging because of structural differences. Therefore, we define \textit{CamerScore}. (Reference video clips: \href{https://www.youtube.com/watch?v=7p0J9wVGwZ0&list=PLZbXA4lyCtqo4dbr3bPiK8-pxWB6wGQHL}{top row},\href{https://youtu.be/Kb2WClrbrAc?si=rhqRk5TjMnHIKSbs}{bottom row}).}
    \label{fig:colmap_opticalflow}
\end{figure*}

Figure~\ref{fig:colmap_opticalflow} illustrates this limitation. In Case 1, the zoom-in effect is achieved through the physical movement of the camera toward the subjects, enabling COLMAP to generate a 3D trajectory for the reference video. In contrast, Case 2 demonstrates a zoom-in effect achieved by altering the focal length, with no actual camera movement. Here, COLMAP fails to compute a trajectory. However, in both cases, homography matrices can still be computed. This highlights the advantage of using homography over COLMAP, making it a more robust and versatile metric for evaluating camera motion.

Consider the optical flow maps generated for the reference and generated videos in both Case-1 and Case-2 (Figure~\ref{fig:colmap_opticalflow}). While the observed motion in both the reference and generated videos is similar, directly comparing their optical flow maps is challenging due to structural differences between the reference scene and the desired scene. Additionally, optical flow often exhibits significant errors near boundaries, making it unreliable for accurately evaluating the observed camera motions.

Considering the different reference video-image pairs, we have a total of $680$ test samples in the dataset. This is comparable to other works in image and video synthesis, such as~\cite{ruiz2023dreambooth}

\section{Qualitative Results}
\begin{figure*}
    \centering
    \includegraphics[width=1\textwidth]{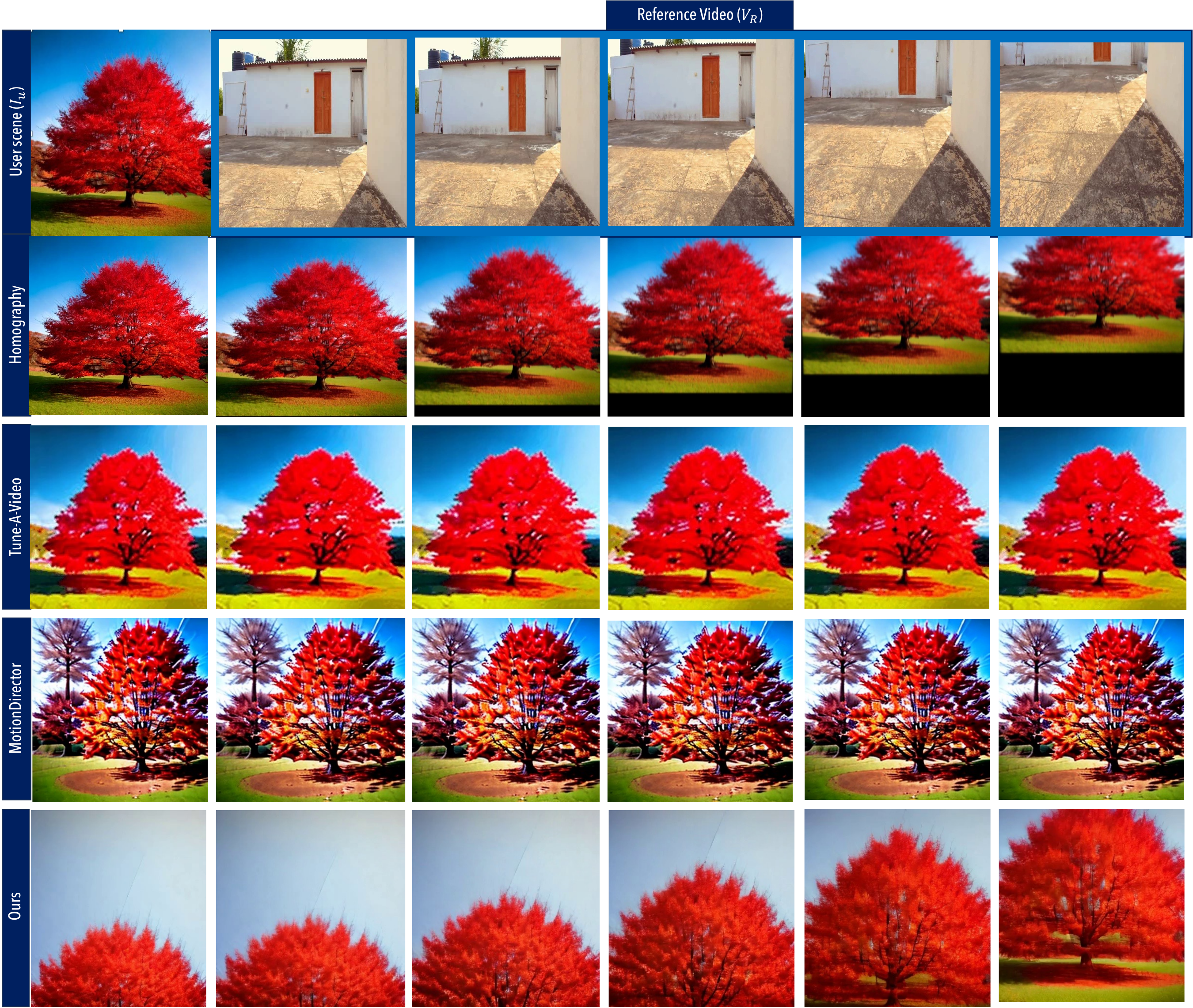}
    \caption{The first row shows the user-provided scene and the reference video (\href{https://www.youtube.com/@MOVIECLIPS}{source}) containing camera motion. In contrast to the baselines (Tune-A-Video and MotionDirector), our method more effectively captures the underlying camera motion from the reference video while also better preserving the details of the user-provided scene.}
    \label{fig:red_tree}
\end{figure*}

\begin{figure*}
    \centering
    \includegraphics[width=1\textwidth]{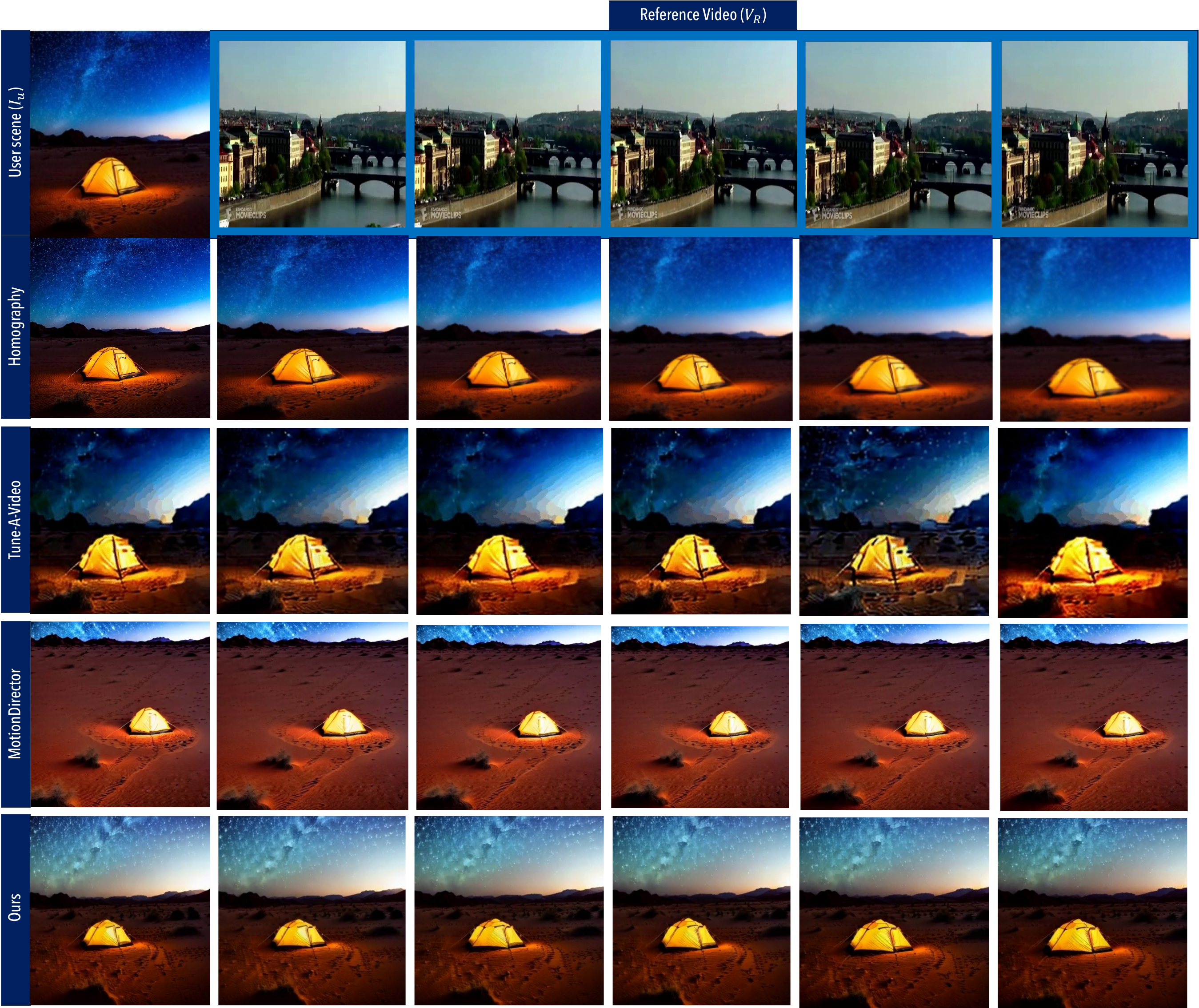}
    \caption{The first row shows the user-provided scene and the reference video (\href{https://www.youtube.com/@MOVIECLIPS}{source}) containing camera motion. In contrast to the baselines (Tune-A-Video and MotionDirector), our method more effectively captures the underlying camera motion from the reference video while also better preserving the details of the user-provided scene.}
    \label{fig:zoom_in_ref6}
\end{figure*}

\begin{figure*}
    \centering
    \includegraphics[width=1\textwidth]{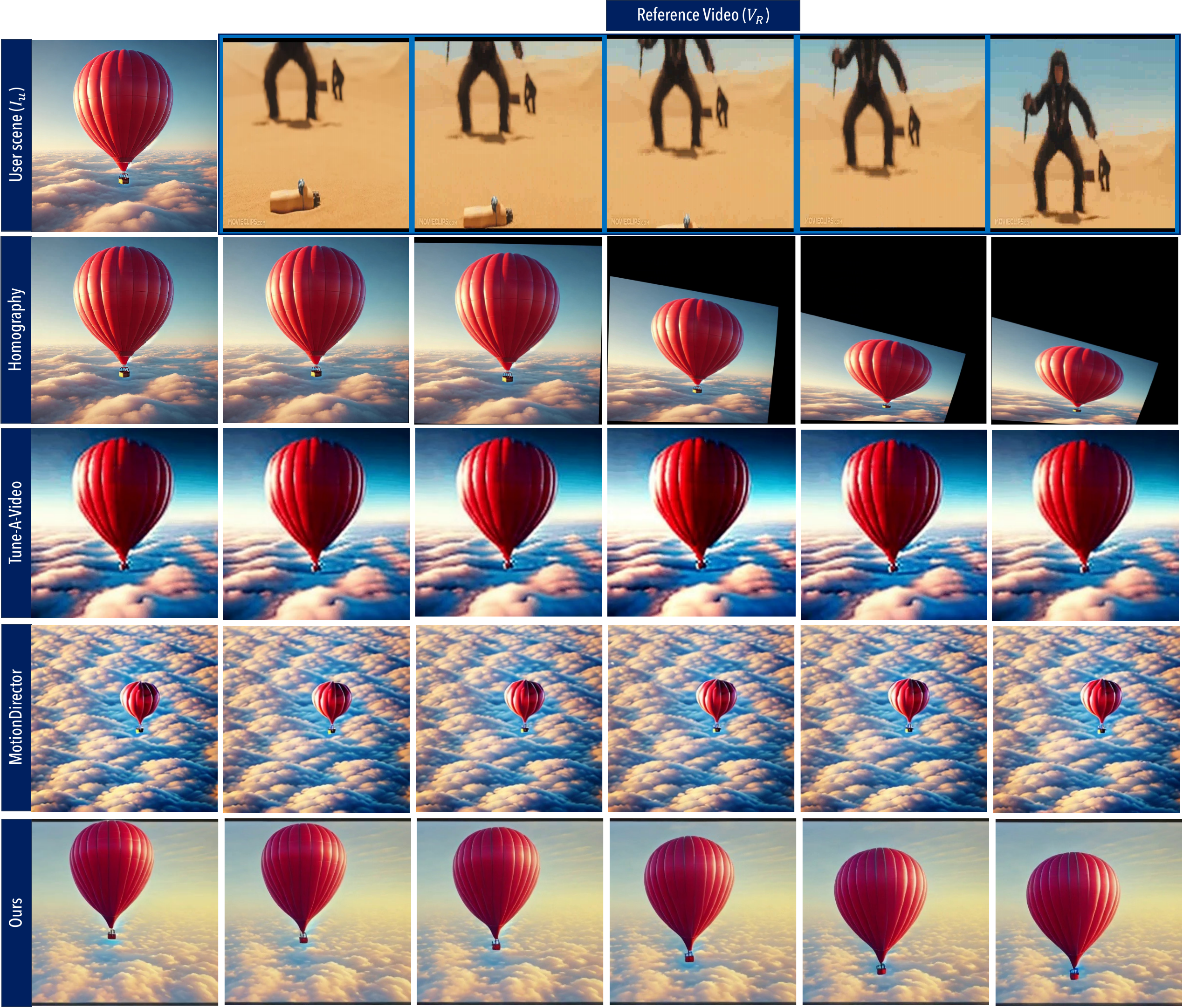}
    \caption{The first row shows the inputs, i.e., the user-provided scene and the reference video (\href{https://www.youtube.com/@MOVIECLIPS}{source}) containing camera motion. In contrast to the baselines, the video output generated by our method more effectively captures the underlying camera motion from the reference video while also better preserving the details of the user-provided scene.}
    \label{fig:tilt_ref7_balloon}
\end{figure*}

\begin{figure*}
    \centering
    \includegraphics[width=1\linewidth]{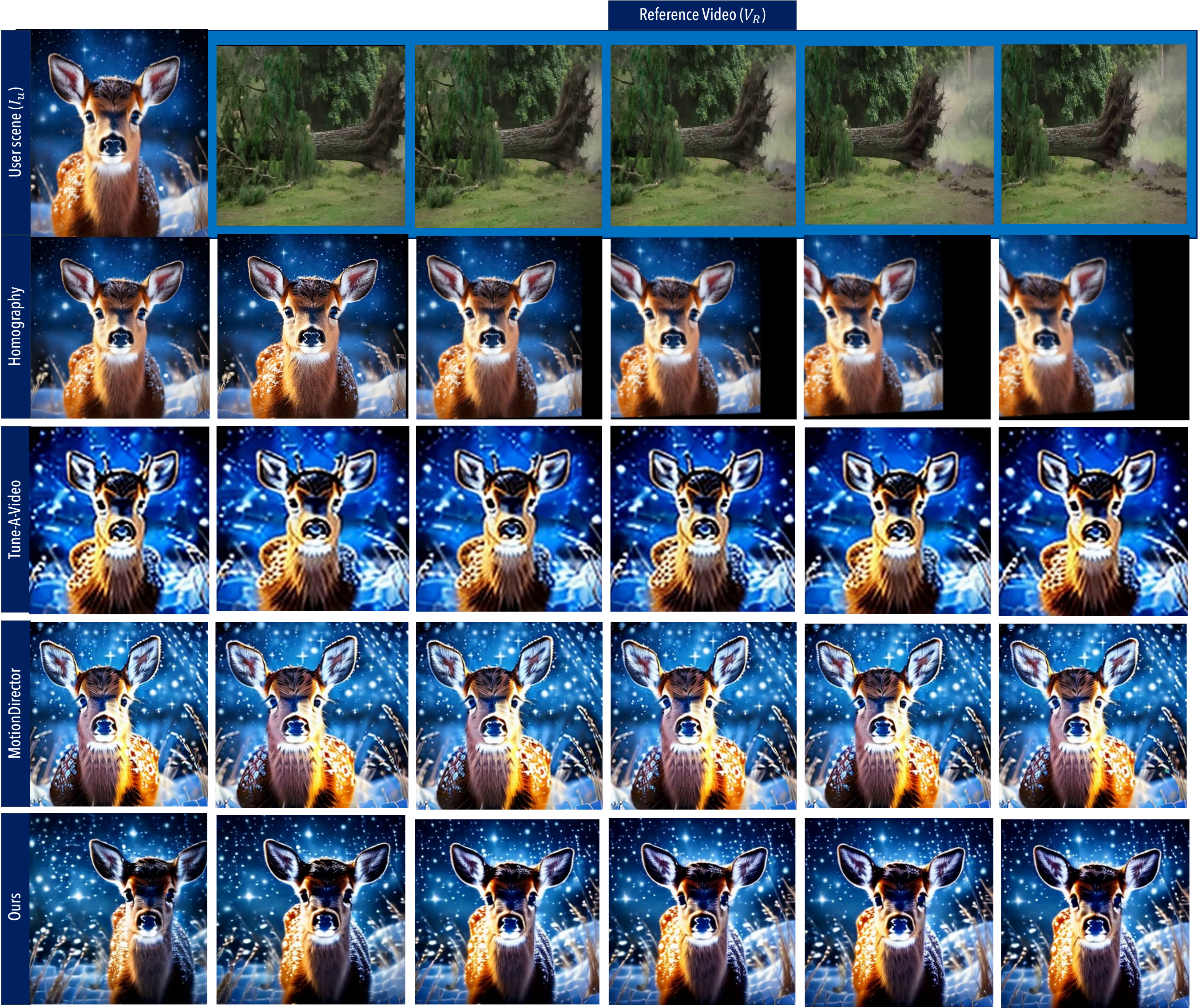}
    \caption{The first row shows the inputs, i.e. the user-provided scene and the reference video (\href{https://www.youtube.com/@MOVIECLIPS}{source}) containing camera motion. In contrast to the baselines, the video output generated by our method more effectively captures the underlying camera motion from the reference video while also better preserving the details of the user-provided scene.}
    \label{fig:pan_ref2_reindeer_baby}
\end{figure*}

\begin{figure*}
    \centering
    \includegraphics[width=1\linewidth]{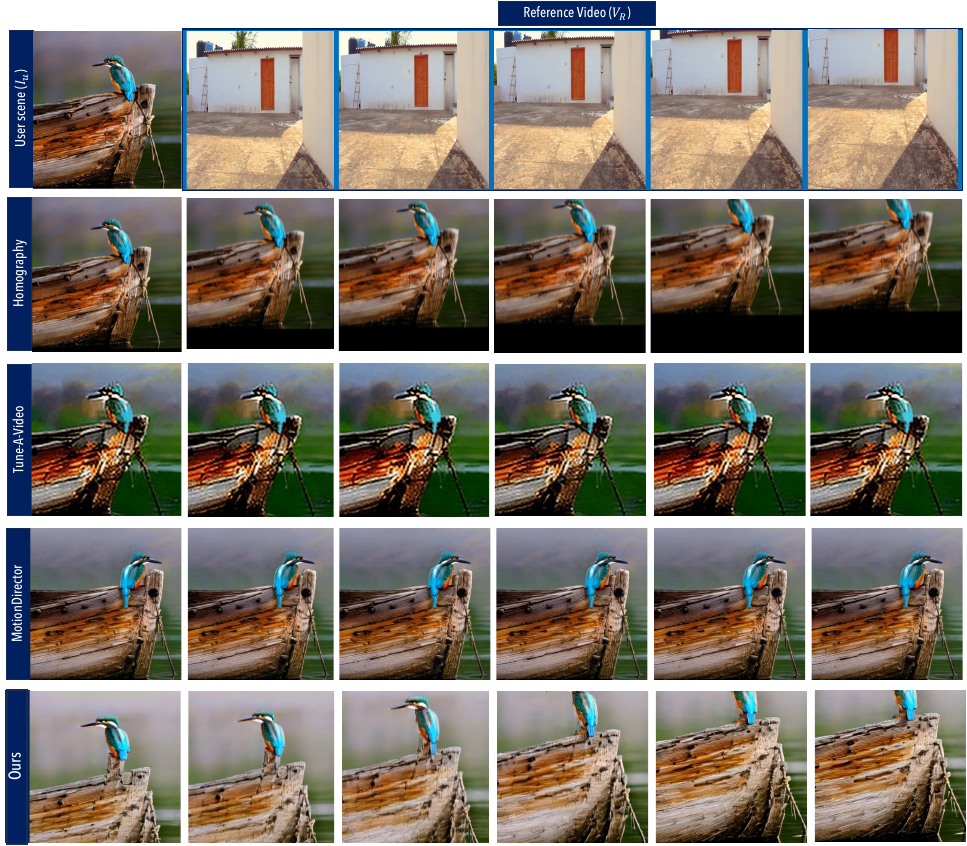}
    \caption{The first row shows the inputs, i.e. the user-provided scene and the reference video (\href{https://www.youtube.com/@MOVIECLIPS}{source}) containing camera motion. In contrast to the baselines, the video output generated by our method more effectively captures the underlying camera motion from the reference video while also better preserving the details of the user-provided scene.}
    \label{fig:kingfisher_tilt_ref10}
\end{figure*}

\begin{figure*}
    \centering
    \includegraphics[width=1\textwidth]{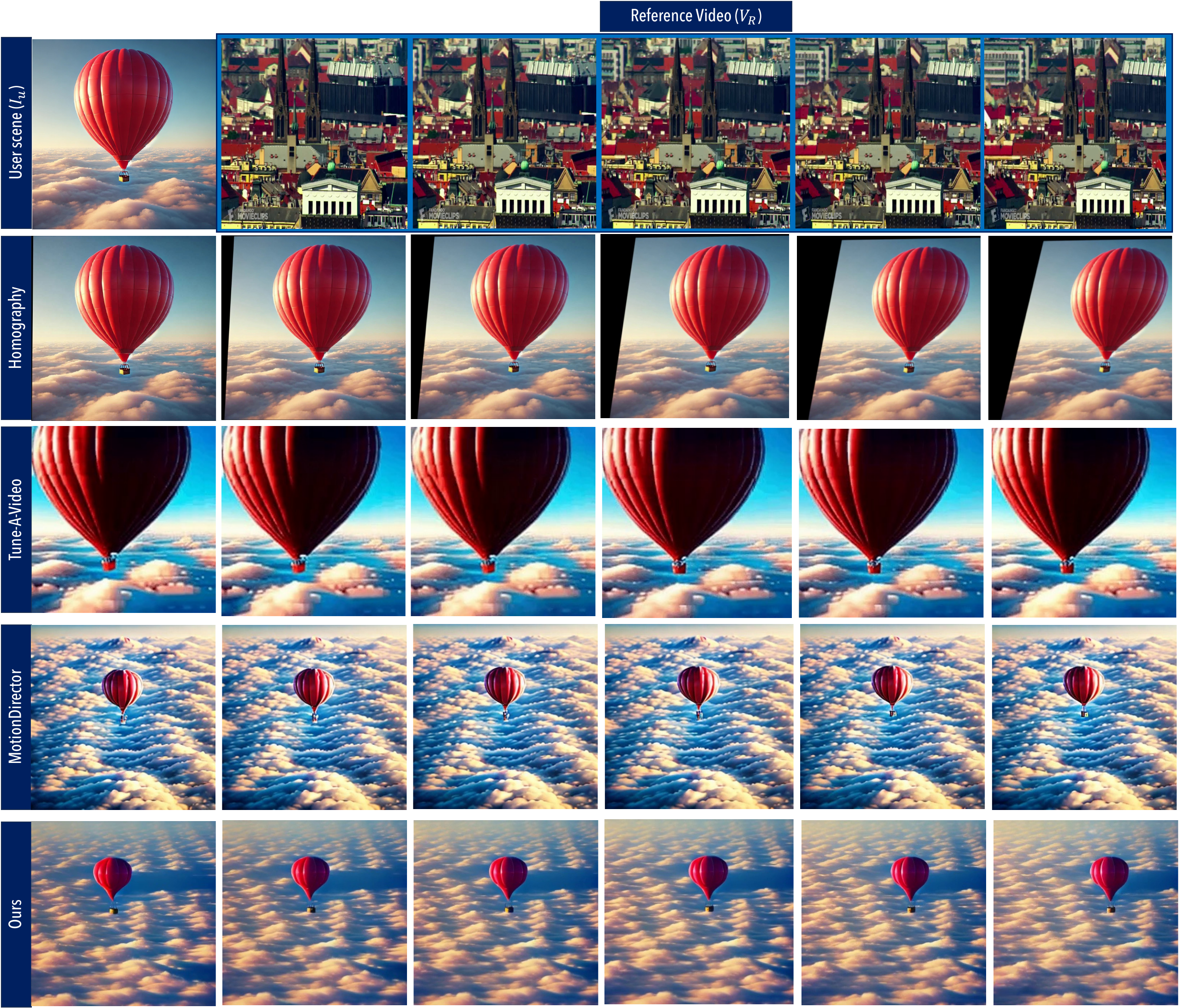}
    \caption{The first row shows the inputs, i.e., the user-provided scene and the reference video (\href{https://www.youtube.com/@MOVIECLIPS}{source}) containing camera motion. In contrast to the baselines,  the video output generated by our method more effectively captures the underlying camera motion from the reference video while also better preserving the details of the user-provided scene.}
    \label{fig:pan_ref1}
\end{figure*}

\begin{figure*}
    \centering
    \includegraphics[width=1\textwidth]{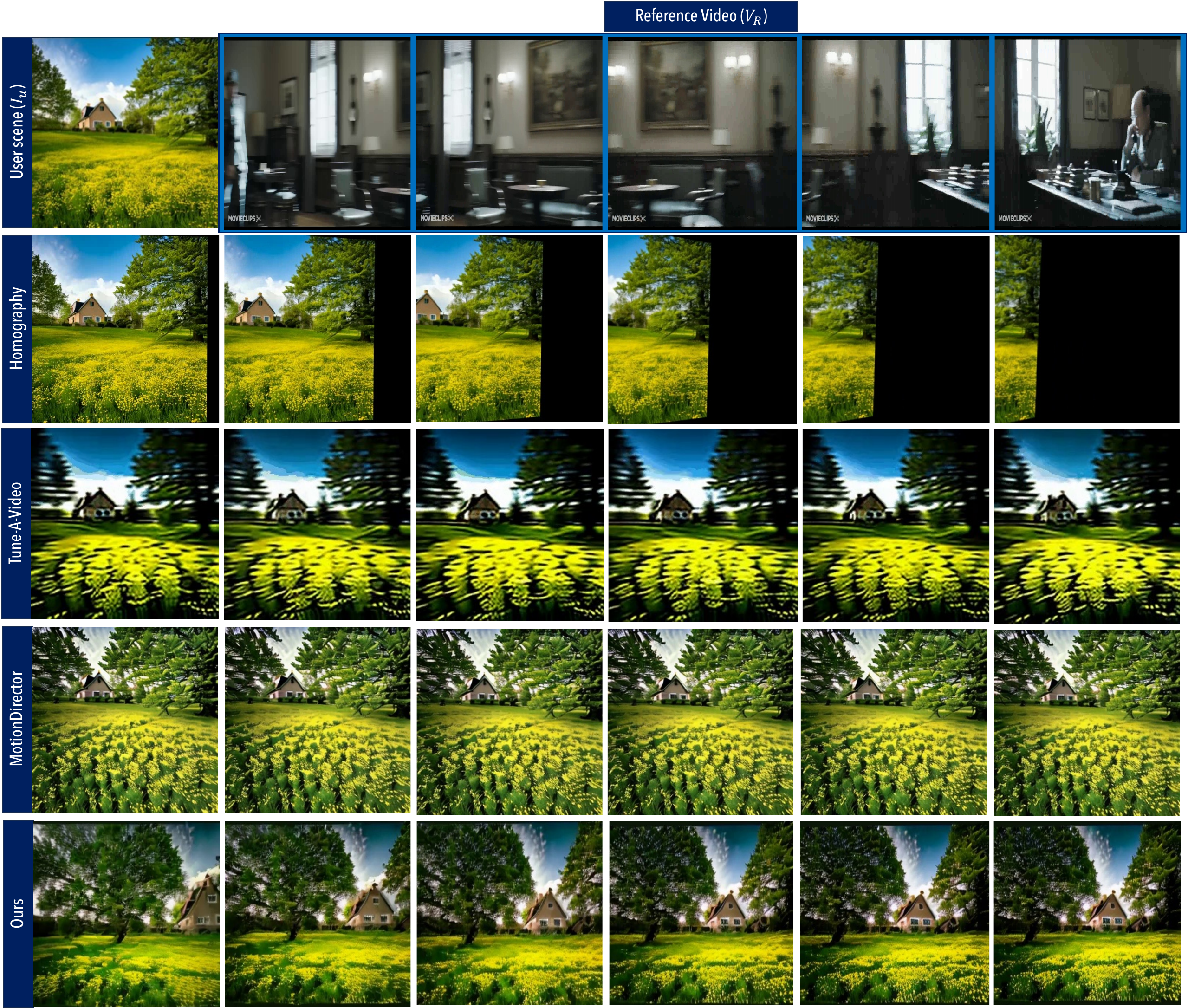}
    \caption{The first row shows the inputs, i.e., the user-provided scene and the reference video (\href{https://www.youtube.com/@MOVIECLIPS}{source}) containing camera motion. In contrast to the baselines, the video output generated by our method more effectively captures the underlying camera motion from the reference video while also better preserving the details of the user-provided scene.}
    \label{fig:pan_ref3_house}
\end{figure*}

We showcase some qualitative results obtained using our method in Fig.\ref{fig:red_tree}-\ref{fig:pan_ref3_house}. In each example shown, we show the video generated using our method as well as baselines. The same video backbone (zeroscope~\cite{zeroscopev2}) has been used in our method and MotionDirector~\cite{zhao2025motiondirector} to generate the video output observed. Stable Diffusion v2~\cite{rombach2022high}~has been used as the image generation backbone for generating Tune-A-Video~\cite{wu2023tune} outputs. \textit{The supplementary video shows additional results with different camera motions.}   


\end{appendices}

\bibliography{sn-bibliography}

@String(ICCV= {Int. Conf. Comput. Vis.})

@String(ECCV= {Eur. Conf. Comput. Vis.})

@String(TOG= {ACM Trans. Graph.})

@String(AAAI = {AAAI})

@String(ICCV  = {ICCV})

@String(ECCV  = {ECCV})

@String(TOG   = {ACM TOG})

@article{guo2023animatediff,
  title={Animatediff: Animate your personalized text-to-image diffusion models without specific tuning},
  author={Guo, Yuwei and Yang, Ceyuan and Rao, Anyi and Liang, Zhengyang and Wang, Yaohui and Qiao, Yu and Agrawala, Maneesh and Lin, Dahua and Dai, Bo},
  journal={arXiv preprint arXiv:2307.04725},
  year={2023}
}

@article{hu2024motionmaster,
  title={MotionMaster: Training-free Camera Motion Transfer For Video Generation},
  author={Hu, Teng and Zhang, Jiangning and Yi, Ran and Wang, Yating and Huang, Hongrui and Weng, Jieyu and Wang, Yabiao and Ma, Lizhuang},
  journal={arXiv preprint arXiv:2404.15789},
  year={2024}
}

@article{hart1988development,
  title={{Development of NASA-TLX (Task Load Index): Results of Empirical and Theoretical Research}},
  author={Hart, SG},
  journal={Human Mental Workload/Elsevier},
  year={1988},
  doi={10.1016/s0166-4115}
}

@article{brooke1996sus,
  title={{SUS: A Quick and Dirty Usability Scale}},
  author={Brooke, J},
  journal={Usability Evaluation in Industry},
  year={1996},
  doi={10.1080/10447310802205776}
}

@misc{google_flow_2025,
  author = {{Google}},
  title = {Flow: AI filmmaking tool designed for Veo},
  year = {2025},
  url = {https://blog.google/technology/ai/google-flow-veo-ai-filmmaking-tool/},
  note = {Accessed: 2025-09-12}
}

@misc{google_veo3_2025,
  author = {{Google DeepMind}},
  title = {Veo 3: State-of-the-art video generation model},
  year = {2025},
  url = {https://deepmind.google/models/veo/},
  note = {Accessed: 2025-09-12}
}

@inproceedings{zhao2025motiondirector,
  author    = {Zhao, Rui and Gu, Yuchao and Wu, Jay Zhangjie and Zhang, David Junhao and Liu, Jia-Wei and Wu, Weijia and Keppo, Jussi and Shou, Mike Zheng},
  title     = {MotionDirector: Motion Customization of Text-to-Video Diffusion Models},
  booktitle = {European Conference on Computer Vision},
  year      = {2025},
  pages     = {273--290},
  publisher = {Springer},
  editor    = {ECCV 2025 Organizing Committee},  
}

@inproceedings{wu2023tune,
  author    = {Wu, Jay Zhangjie and Ge, Yixiao and Wang, Xintao and
               Lei, Stan Weixian and Gu, Yuchao and Shi, Yufei and
               Hsu, Wynne and Shan, Ying and Qie, Xiaohu and Shou, Mike Zheng},
  title     = {Tune-A-Video: One-Shot Tuning of Image Diffusion Models for Text-to-Video Generation},
  booktitle = {Proceedings of the IEEE/CVF International Conference on Computer Vision},
  year      = {2023},
  month     = {October},
  pages     = {7623--7633},
  publisher = {IEEE},
  editor    = {IEEE Computer Society},
}

@article{he2024cameractrl,
  title={Cameractrl: Enabling camera control for text-to-video generation},
  author={He, Hao and Xu, Yinghao and Guo, Yuwei and Wetzstein, Gordon and Dai, Bo and Li, Hongsheng and Yang, Ceyuan},
  journal={arXiv preprint arXiv:2404.02101},
  year={2024}
}

@inproceedings{ruiz2023dreambooth,
  author    = {Ruiz, Nataniel and Li, Yuanzhen and Jampani, Varun and
               Pritch, Yael and Rubinstein, Michael and Aberman, Kfir},
  title     = {DreamBooth: Fine Tuning Text-to-Image Diffusion Models for Subject-Driven Generation},
  booktitle = {Proceedings of the IEEE/CVF Conference on Computer Vision and Pattern Recognition},
  year      = {2023},
  month     = {June},
  pages     = {22500--22510},
  publisher = {IEEE},
  editor    = {IEEE Computer Society}, 
}

@article{blattmann2023stable,
  title={Stable video diffusion: Scaling latent video diffusion models to large datasets},
  author={Blattmann, Andreas and Dockhorn, Tim and Kulal, Sumith and Mendelevitch, Daniel and Kilian, Maciej and Lorenz, Dominik and Levi, Yam and English, Zion and Voleti, Vikram and Letts, Adam and others},
  journal={arXiv preprint arXiv:2311.15127},
  year={2023}
}

@article{wang2023modelscope,
  title={Modelscope text-to-video technical report},
  author={Wang, Jiuniu and Yuan, Hangjie and Chen, Dayou and Zhang, Yingya and Wang, Xiang and Zhang, Shiwei},
  journal={arXiv preprint arXiv:2308.06571},
  year={2023}
}

@misc{zeroscopev2,
 title = {Zeroscope},
 url = {https://huggingface.co/cerspense/zeroscope_v2_576w},
 year = 2023
}

@article{polyak2024movie,
  title={Movie gen: A cast of media foundation models},
  author={Polyak, Adam and Zohar, Amit and Brown, Andrew and Tjandra, Andros and Sinha, Animesh and Lee, Ann and Vyas, Apoorv and Shi, Bowen and Ma, Chih-Yao and Chuang, Ching-Yao and others},
  journal={arXiv preprint arXiv:2410.13720},
  year={2024}
}

@article{hu2021lora,
  title={Lora: Low-rank adaptation of large language models},
  author={Hu, Edward J and Shen, Yelong and Wallis, Phillip and Allen-Zhu, Zeyuan and Li, Yuanzhi and Wang, Shean and Wang, Lu and Chen, Weizhu},
  journal={arXiv preprint arXiv:2106.09685},
  year={2021}
}

@inproceedings{zhang2023adding,
  author    = {Zhang, Lvmin and Rao, Anyi and Agrawala, Maneesh},
  title     = {Adding Conditional Control to Text-to-Image Diffusion Models},
  booktitle = {Proceedings of the IEEE/CVF International Conference on Computer Vision},
  year      = {2023},
  month     = {October},
  pages     = {3836--3847},
  publisher = {IEEE},
  editor    = {IEEE Computer Society},  
}

@inproceedings{tewel2023key,
  author    = {Tewel, Yoad and Gal, Rinon and Chechik, Gal and Atzmon, Yuval},
  title     = {Key-Locked Rank One Editing for Text-to-Image Personalization},
  booktitle = {ACM SIGGRAPH 2023 Conference Proceedings},
  year      = {2023},
  pages     = {1--11},
  publisher = {Association for Computing Machinery},
  editor    = {Stephen N. Spencer},     
}

@inproceedings{li2025tuning,
  title={Tuning-free image customization with image and text guidance},
  author={Li, Pengzhi and Nie, Qiang and Chen, Ying and Jiang, Xi and Wu, Kai and Lin, Yuhuan and Liu, Yong and Peng, Jinlong and Wang, Chengjie and Zheng, Feng},
  booktitle={European Conference on Computer Vision},
  year        = {2025},
  pages       = {233--250},
  publisher   = {Springer},   
  editor = {ECCV 2025 Organizing Committee}
}

@article{alaluf2023neural,
  title={A neural space-time representation for text-to-image personalization},
  author={Alaluf, Yuval and Richardson, Elad and Metzer, Gal and Cohen-Or, Daniel},
  journal={ACM Transactions on Graphics (TOG)},
  volume={42},
  number={6},
  pages={1--10},
  year={2023},
  publisher={ACM New York, NY, USA}
}

@inproceedings{chen2024anydoor,
  author    = {Chen, Xi and Huang, Lianghua and Liu, Yu and Shen, Yujun and
               Zhao, Deli and Zhao, Hengshuang},
  title     = {AnyDoor: Zero-shot Object-level Image Customization},
  booktitle = {Proceedings of the IEEE/CVF Conference on Computer Vision and Pattern Recognition},
  year      = {2024},
  month     = {June},
  pages     = {6593--6602},
  publisher = {IEEE},
  editor    = {IEEE Computer Society},
}

@article{yuan2023customnet,
  title={Customnet: Zero-shot object customization with variable-viewpoints in text-to-image diffusion models},
  author={Yuan, Ziyang and Cao, Mingdeng and Wang, Xintao and Qi, Zhongang and Yuan, Chun and Shan, Ying},
  journal={arXiv preprint arXiv:2310.19784},
  year={2023}
}

@article{ma2017pose,
  title={Pose guided person image generation},
  author={Ma, Liqian and Jia, Xu and Sun, Qianru and Schiele, Bernt and Tuytelaars, Tinne and Van Gool, Luc},
  journal={Advances in neural information processing systems},
  volume={30},
  year={2017}
}

@inproceedings{nguyen2024nope,
  author    = {Nguyen, Van Nguyen and Groueix, Thibault and Ponimatkin, Georgy and
               Hu, Yinlin and Marlet, Renaud and Salzmann, Mathieu and Lepetit, Vincent},
  title     = {NOPE: Novel Object Pose Estimation from a Single Image},
  booktitle = {Proceedings of the IEEE/CVF Conference on Computer Vision and Pattern Recognition},
  year      = {2024},
  month     = {June},
  pages     = {17923--17932},
  publisher = {IEEE},
  editor    = {IEEE Computer Society}, 
}

@inproceedings{abbaspour2017practical,
  author    = {Abbaspour Tehrani, Mahdi and Beeler, Thabo and Grundhofer, Anselm},
  title     = {A Practical Method for Fully Automatic Intrinsic Camera Calibration Using Directionally Encoded Light},
  booktitle = {Proceedings of the IEEE Conference on Computer Vision and Pattern Recognition},
  year      = {2017},
  pages     = {1106--1114},          
  publisher = {IEEE},
  editor    = {IEEE Computer Society}, 
}

@article{xu2024sgdm,
  title={SGDM: An Adaptive Style-Guided Diffusion Model for Personalized Text to Image Generation},
  author={Xu, Yifei and Xu, Xiaolong and Gao, Honghao and Xiao, Fu},
  journal={IEEE Transactions on Multimedia},
  year={2024},
  publisher={IEEE}
}

@inproceedings{ruiz2024hyperdreambooth,
  author    = {Ruiz, Nataniel and Li, Yuanzhen and Jampani, Varun and Wei, Wei and Hou, Tingbo and Pritch, Yael and Wadhwa, Neal and Rubinstein, Michael and Aberman, Kfir},
  title     = {HyperDreamBooth: HyperNetworks for Fast Personalization of Text-to-Image Models},
  booktitle = {Proceedings of the IEEE/CVF Conference on Computer Vision and Pattern Recognition},
  year      = {2024},
  month     = {June},
  pages     = {6527--6536},
  publisher = {IEEE},
  editor    = {IEEE Computer Society},
}

@article{zhang2024recapture,
  title={ReCapture: Generative Video Camera Controls for User-Provided Videos using Masked Video Fine-Tuning},
  author={Zhang, David Junhao and Paiss, Roni and Zada, Shiran and Karnad, Nikhil and Jacobs, David E and Pritch, Yael and Mosseri, Inbar and Shou, Mike Zheng and Wadhwa, Neal and Ruiz, Nataniel},
  journal={arXiv preprint arXiv:2411.05003},
  year={2024}
}

@article{imambi2021pytorch,
  title={PyTorch},
  author={Imambi, Sagar and Prakash, Kolla Bhanu and Kanagachidambaresan, GR},
  journal={Programming with TensorFlow: Solution for Edge Computing Applications},
  pages={87--104},
  year={2021},
  publisher={Springer}
}

@article{oquab2023dinov2,
  title={Dinov2: Learning robust visual features without supervision},
  author={Oquab, Maxime and Darcet, Timoth{\'e}e and Moutakanni, Th{\'e}o and Vo, Huy and Szafraniec, Marc and Khalidov, Vasil and Fernandez, Pierre and Haziza, Daniel and Massa, Francisco and El-Nouby, Alaaeldin and others},
  journal={arXiv preprint arXiv:2304.07193},
  year={2023}
}

@article{xu2021videoclip,
  title={Videoclip: Contrastive pre-training for zero-shot video-text understanding},
  author={Xu, Hu and Ghosh, Gargi and Huang, Po-Yao and Okhonko, Dmytro and Aghajanyan, Armen and Metze, Florian and Zettlemoyer, Luke and Feichtenhofer, Christoph},
  journal={arXiv preprint arXiv:2109.14084},
  year={2021}
}

@inproceedings{jeong2024vmc,
  author    = {Jeong, Hyeonho and Park, Geon Yeong and Ye, Jong Chul},
  title     = {VMC: Video Motion Customization Using Temporal Attention Adaption for Text-to-Video Diffusion Models},
  booktitle = {Proceedings of the IEEE/CVF Conference on Computer Vision and Pattern Recognition},
  year      = {2024},
  month     = {June},
  pages     = {9212--9221},
  publisher = {IEEE},
  editor    = {IEEE Computer Society},  
}

@article{ren2024customize,
  title={Customize-a-video: One-shot motion customization of text-to-video diffusion models},
  author={Ren, Yixuan and Zhou, Yang and Yang, Jimei and Shi, Jing and Liu, Difan and Liu, Feng and Kwon, Mingi and Shrivastava, Abhinav},
  journal={arXiv preprint arXiv:2402.14780},
  year={2024}
}

@article{zhang2023motioncrafter,
  title={Motioncrafter: One-shot motion customization of diffusion models},
  author={Zhang, Yuxin and Tang, Fan and Huang, Nisha and Huang, Haibin and Ma, Chongyang and Dong, Weiming and Xu, Changsheng},
  journal={arXiv preprint arXiv:2312.05288},
  year={2023}
}

@article{materzynska2023customizing,
  title={Customizing motion in text-to-video diffusion models},
  author={Materzynska, Joanna and Sivic, Josef and Shechtman, Eli and Torralba, Antonio and Zhang, Richard and Russell, Bryan},
  journal={arXiv preprint arXiv:2312.04966},
  year={2023}
}

@article{wang2024customvideo,
  title={Customvideo: Customizing text-to-video generation with multiple subjects},
  author={Wang, Zhao and Li, Aoxue and Zhu, Lingting and Guo, Yong and Dou, Qi and Li, Zhenguo},
  journal={arXiv preprint arXiv:2401.09962},
  year={2024}
}

@article{shi2023zero123++,
  title={Zero123++: a single image to consistent multi-view diffusion base model},
  author={Shi, Ruoxi and Chen, Hansheng and Zhang, Zhuoyang and Liu, Minghua and Xu, Chao and Wei, Xinyue and Chen, Linghao and Zeng, Chong and Su, Hao},
  journal={arXiv preprint arXiv:2310.15110},
  year={2023}
}

@article{liu2024one,
  title={One-2-3-45: Any single image to 3d mesh in 45 seconds without per-shape optimization},
  author={Liu, Minghua and Xu, Chao and Jin, Haian and Chen, Linghao and Varma T, Mukund and Xu, Zexiang and Su, Hao},
  journal={Advances in Neural Information Processing Systems},
  volume={36},
  year={2024}
}

@article{kothandaraman2024text,
  title={Text Prompting for Multi-Concept Video Customization by Autoregressive Generation},
  author={Kothandaraman, Divya and Sohn, Kihyuk and Villegas, Ruben and Voigtlaender, Paul and Manocha, Dinesh and Babaeizadeh, Mohammad},
  journal={arXiv preprint arXiv:2405.13951},
  year={2024}
}

@article{ho2022classifier,
  title={Classifier-free diffusion guidance},
  author={Ho, Jonathan and Salimans, Tim},
  journal={arXiv preprint arXiv:2207.12598},
  year={2022}
}

@book{szeliski2022computer,
  title={Computer vision: algorithms and applications},
  author={Szeliski, Richard},
  year={2022},
  publisher={Springer Nature}
}

@article{beauchemin1995computation,
  title={The computation of optical flow},
  author={Beauchemin, Steven S. and Barron, John L.},
  journal={ACM computing surveys (CSUR)},
  volume={27},
  number={3},
  pages={433--466},
  year={1995},
  publisher={ACM New York, NY, USA}
}

@inproceedings{schoenberger2016sfm,
  author    = {Sch{\"o}nberger, Johannes Lutz and Frahm, Jan-Michael},
  title     = {Structure-from-Motion Revisited},
  booktitle = {Proceedings of the IEEE Conference on Computer Vision and Pattern Recognition},
  year      = {2016},
  pages     = {4104--4113},
  publisher = {IEEE},
  editor    = {IEEE Computer Society},  
}

@article{jiang2024cinematographic,
  author   = {Jiang, Hongda and Wang, Xi and Christie, Marc and Liu, Libin and Chen, Baoquan},
  title    = {Cinematographic Camera Diffusion Model},
  journal  = {Computer Graphics Forum},
  year     = {2024},
  volume   = {43},
  number   = {2},
  pages    = {e15055},
  publisher= {Wiley},
}

@article{asadi2024does,
  title={Does combining parameter-efficient modules improve few-shot transfer accuracy?},
  author={Asadi, Nader and Beitollahi, Mahdi and Khalil, Yasser and Li, Yinchuan and Zhang, Guojun and Chen, Xi},
  journal={arXiv preprint arXiv:2402.15414},
  year={2024}
}

@article{kothandaraman2023hawki,
  title={HawkI: Homography \& Mutual Information Guidance for 3D-free Single Image to Aerial View},
  author={Kothandaraman, Divya and Zhou, Tianyi and Lin, Ming and Manocha, Dinesh},
  journal={arXiv preprint arXiv:2311.15478},
  year={2023}
}

@article{zhuang2024copra,
  title={CopRA: A Progressive LoRA Training Strategy},
  author={Zhuang, Zhan and Wang, Xiequn and Zhang, Yulong and Li, Wei and Zhang, Yu and Wei, Ying},
  journal={arXiv preprint arXiv:2410.22911},
  year={2024}
}

@misc{videographertips,
 title = {Videographer Tips: How to Direct Your Audience with Camera Movement},
 url = {https://hayotfilms.com/blog/videographer-tips-camera-movement/},
 year = 2024
}

@misc{rubberducksmovementstory,
 title = {How to Use Movement to Tell a Story in Your Videography},
 url = {https://rubberduckers.co.uk/how-to-use-movement-to-tell-a-story-in-your-videography/},
 year = 2024
}

@misc{cameramotiondefn,
 title = {Camera Movement Terms: Everything You Need To Know},
 url = {https://www.nfi.edu/camera-movement-terms/},
 year = 2024
}

@misc{productionstory,
 title = {The Challenges of Shooting Photography and Videography},
 url = {https://www.adorama.com/alc/shooting-photography-and-videography/amp/
},
 year = 2025
}

@misc{moviegen_meta,
 title = {Movie Gen: A Cast of Media Foundation Models},
 url = {https://ai.meta.com/research/publications/movie-gen-a-cast-of-media-foundation-models/
},
 year = 2024
}

@inproceedings{hu2024comd,
  author    = {Hu, Teng and Zhang, Jiangning and Yi, Ran and Wang, Yating and
               Weng, Jieyu and Huang, Hongrui and Wang, Yabiao and Ma, Lizhuang},
  title     = {COMD: Training-free Video Motion Transfer With Camera-Object Motion Disentanglement},
  booktitle = {Proceedings of the 32nd ACM International Conference on Multimedia},
  year      = {2024},
  pages     = {3459--3468},
  publisher = {Association for Computing Machinery}, 
  editor    = {ACM MM 2024 Organizing Committee},
}

@article{veo2,
  author  = {Veo-Team and Agrim Gupta and Ali Razavi and Andeep Toor and
             Ankush Gupta and Dumitru Erhan and Eleni Shaw and Eric Lau and
             Frank Belletti and Gabe Barth-Maron and Gregory Shaw and
             Hakan Erdogan and Hakim Sidahmed and Henna Nandwani and
             Hernan Moraldo and Hyunjik Kim and Irina Blok and Jeff Donahue and
             José Lezama and Kory Mathewson and Kurtis David and
             Matthieu Kim Lorrain and Marc van Zee and Medhini Narasimhan and
             Miaosen Wang and Mohammad Babaeizadeh and Nelly Papalampidi and
             Nick Pezzotti and Nilpa Jha and Parker Barnes and
             Pieter-Jan Kindermans and Rachel Hornung and Ruben Villegas and
             Ryan Poplin and Salah Zaiem and Sander Dieleman and
             Sayna Ebrahimi and Scott Wisdom and Serena Zhang and
             Shlomi Fruchter and Signe Nørly and Weizhe Hua and
             Xinchen Yan and Yuqing Du and Yutian Chen},
  title   = {Veo 2},
  journal = {DeepMind Technical Report},  
  year    = {2024},
  url     = {https://deepmind.google/technologies/veo/veo-2/},
}

@inproceedings{shi2024instantbooth,
  author    = {Shi, Jing and Xiong, Wei and Lin, Zhe and Jung, Hyun Joon},
  title     = {InstantBooth: Personalized Text-to-Image Generation without Test-Time Finetuning},
  booktitle = {Proceedings of the IEEE/CVF Conference on Computer Vision and Pattern Recognition},
  year      = {2024},
  month     = {June},
  pages     = {8543--8552},
  publisher = {IEEE},
  editor    = {IEEE Computer Society}, 
}

@inproceedings{li2024photomaker,
  author    = {Li, Zhen and Cao, Mingdeng and Wang, Xintao and Qi, Zhongang and Cheng, Ming-Ming and Shan, Ying},
  title     = {PhotoMaker: Customizing Realistic Human Photos via Stacked ID Embedding},
  booktitle = {Proceedings of the IEEE/CVF Conference on Computer Vision and Pattern Recognition},
  year      = {2024},
  pages     = {8640--8650},
  publisher = {IEEE},
  editor    = {IEEE Computer Society}, 
}

@inproceedings{xing2024dynamicrafter,
  author    = {Xing, Jinbo and Xia, Menghan and Zhang, Yong and Chen, Haoxin and
               Yu, Wangbo and Liu, Hanyuan and Liu, Gongye and Wang, Xintao and
               Shan, Ying and Wong, Tien-Tsin},
  title     = {Dynamicrafter: Animating Open-Domain Images with Video Diffusion Priors},
  booktitle = {European Conference on Computer Vision},
  year      = {2024},
  pages     = {399--417},
  publisher = {Springer},
  editor    = {ECCV 2024 Organizing Committee},
}

@article{wu2024motionbooth,
  title={Motionbooth: Motion-aware customized text-to-video generation},
  author={Wu, Jianzong and Li, Xiangtai and Zeng, Yanhong and Zhang, Jiangning and Zhou, Qianyu and Li, Yining and Tong, Yunhai and Chen, Kai},
  journal={arXiv preprint arXiv:2406.17758},
  year={2024}
}

@inproceedings{ma2024follow,
  title={Follow your pose: Pose-guided text-to-video generation using pose-free videos},
  author={Ma, Yue and He, Yingqing and Cun, Xiaodong and Wang, Xintao and Chen, Siran and Li, Xiu and Chen, Qifeng},
  booktitle={Proceedings of the AAAI Conference on Artificial Intelligence},
  year      = {2024},
  pages     = {4117--4125},
  volume    = {38},                 
  publisher = {AAAI Press},    
  editor = {AAAI-24 Organizing Committee},

}

@inproceedings{zhang2024pia,
  author    = {Zhang, Yiming and Xing, Zhening and Zeng, Yanhong and Fang, Youqing and Chen, Kai},
  title     = {Pia: Your Personalized Image Animator via Plug-and-Play Modules in Text-to-Image Models},
  booktitle = {Proceedings of the IEEE/CVF Conference on Computer Vision and Pattern Recognition},
  year      = {2024},
  pages     = {7747--7756},
  publisher = {IEEE},
  editor    = {IEEE Computer Society}
}

@article{totlani2023evolution,
  title={The evolution of generative AI: Implications for the media and film industry},
  author={Totlani, Ketan},
  journal={International Journal for Research in Applied Science and Engineering Technology},
  volume={11},
  number={10},
  pages={973--980},
  year={2023}
}

@article{he2024imagine,
  title={Imagine yourself: Tuning-free personalized image generation},
  author={He, Zecheng and Sun, Bo and Juefei-Xu, Felix and Ma, Haoyu and Ramchandani, Ankit and Cheung, Vincent and Shah, Siddharth and Kalia, Anmol and Subramanyam, Harihar and Zareian, Alireza and others},
  journal={arXiv preprint arXiv:2409.13346},
  year={2024}
}

@article{dhariwal2021diffusion,
  title={Diffusion models beat gans on image synthesis},
  author={Dhariwal, Prafulla and Nichol, Alexander},
  journal={Advances in neural information processing systems},
  volume={34},
  pages={8780--8794},
  year={2021}
}

@article{derpanis2010overview,
  title={Overview of the RANSAC Algorithm},
  author={Derpanis, Konstantinos G},
  journal={Image Rochester NY},
  volume={4},
  number={1},
  pages={2--3},
  year={2010}
}

@article{lowe2004sift,
  title={Sift-the scale invariant feature transform},
  author={Lowe, G},
  journal={Int. J},
  volume={2},
  number={91-110},
  pages={2},
  year={2004}
}

@inproceedings{rombach2022high,
  author    = {Rombach, Robin and Blattmann, Andreas and Lorenz, Dominik and
               Esser, Patrick and Ommer, Bj{\"o}rn},
  title     = {High-Resolution Image Synthesis with Latent Diffusion Models},
  booktitle = {Proceedings of the IEEE/CVF Conference on Computer Vision and Pattern Recognition},
  year      = {2022},
  month     = {June},
  pages     = {10684--10695},
  publisher = {IEEE},
  editor    = {IEEE Computer Society}, 
}

@String{Computing = "Computing" }

@String{Computer = "{IEEE} Computer" }

@String{Springer = "Springer-Verlag" }

@ArtifactSoftware{R,
    title = {R: A Language and Environment for Statistical Computing},
    author = {{R Core Team}},
    organization = {R Foundation for Statistical Computing},
    address = {Vienna, Austria},
    year = {2019},
    url = {https://www.R-project.org/},
}

@article{photoshop2021adobe,
  title={Adobe{\textregistered}},
  author={Photoshop, Adobe and Premiere, Adobe and Pro, CS and CS, Adobe Bridge},
  journal={Praha: Adobe, c2019 [cit. 2019-05-21]. Dostupn{\'e} z: https://www. adobe. com/cz/products/photoshop. html},
  year={2021}
}

@inproceedings{tilekbay2024expressedit,
  author    = {Tilekbay, Bekzat and Yang, Saelyne and Lewkowicz, Michal Adam and
               Suryapranata, Alex and Kim, Juho},
  title     = {ExpressEdit: Video Editing with Natural Language and Sketching},
  booktitle = {Proceedings of the 29th International Conference on Intelligent User Interfaces},
  year      = {2024},
  pages     = {515--536},
  publisher = {Association for Computing Machinery}, 
  editor    = {IUI 2024 Organizing Committee},       
}

@inproceedings{wang2025vggt,
  author    = {Wang, Jianyuan and Chen, Minghao and Karaev, Nikita and
               Vedaldi, Andrea and Rupprecht, Christian and Novotny, David},
  title     = {VGGT: Visual Geometry Grounded Transformer},
  booktitle = {Proceedings of the IEEE/CVF Conference on Computer Vision and Pattern Recognition},
  year      = {2025},
  pages     = {5294--5306},
  publisher = {IEEE},
  editor    = {IEEE Computer Society}, 
}

@Inbook{ref1,
editor="Furht, Borko",
title="Camera Motions",
bookTitle="Encyclopedia of Multimedia",
year="2008",
publisher="Springer US",
address="Boston, MA",
pages="57--58",
isbn="978-0-387-78414-4",
doi="10.1007/978-0-387-78414-4_10",
url="https://doi.org/10.1007/978-0-387-78414-4_10"
}

@incollection{van2013generative,
  title={Generative artificial intelligence},
  author={van der Zant, Tijn and Kouw, Matthijs and Schomaker, Lambert},
  booktitle={Philosophy and theory of artificial intelligence},
  editor    = {M{\"u}ller, Vincent C.},
  year      = {2013},
  publisher = {Springer},
  address   = {Berlin, Heidelberg},
  pages     = {107--120},
}

@article{assembly2015sustainable,
  title={Sustainable development goals},
  author={Assembly, General and others},
  journal={SDGs transform our world},
  volume={2030},
  number={10.1186},
  year={2015}
}

@article{ma2025video,
  title={Video diffusion generation: comprehensive review and open problems},
  author={Ma, Wenping and Yang, Xiaoting and Jiao, Licheng and Li, Lingling and Liu, Xu and Liu, Fang and Chen, Puhua and Yang, Yuting and Ma, Mengru and Sun, Long and others},
  journal={Artificial Intelligence Review},
  volume={58},
  number={11},
  pages={338},
  year={2025},
  publisher={Springer}
}

@article{yu2025trajectorycrafter,
  title={Trajectorycrafter: Redirecting camera trajectory for monocular videos via diffusion models},
  author={YU, Mark and Hu, Wenbo and Xing, Jinbo and Shan, Ying},
  journal={arXiv preprint arXiv:2503.05638},
  year={2025}
}

@inproceedings{luo2025camclonemaster,
  author    = {Luo, Yawen and Shi, Xiaoyu and Bai, Jianhong and Xia, Menghan and Xue, Tianfan and Wang, Xintao and Wan, Pengfei and Zhang, Di and Gai, Kun},
  title     = {CamCloneMaster: Enabling Reference-based Camera Control for Video Generation},
  booktitle = {Proceedings of the SIGGRAPH Asia 2025 Conference Papers},
  year      = {2025},
  pages     = {1--10},
  publisher = {Association for Computing Machinery},
  editor    = {SIGGRAPH Asia 2025 Papers Chair},  
}

@inproceedings{bai2025recammaster,
  author    = {Bai, Jianhong and Xia, Menghan and Fu, Xiao and Wang, Xintao and Mu, Lianrui and Cao, Jinwen and Liu, Zuozhu and Hu, Haoji and Bai, Xiang and Wan, Pengfei and Zhang, Di},
  title     = {ReCamMaster: Camera-Controlled Generative Rendering from a Single Video},
  booktitle = {Proceedings of the IEEE/CVF International Conference on Computer Vision (ICCV)},
  year      = {2025},
  month     = {October},
  pages     = {14834--14844},
  publisher = {IEEE},
  editor    = {IEEE Computer Society},  
}

@inproceedings{wang2025akira,
  author    = {Wang, Xi and Courant, Robin and Christie, Marc and Kalogeiton, Vicky},
  title     = {AKiRa: Augmentation Kit on Rays for Optical Video Generation},
  booktitle = {Proceedings of the IEEE/CVF Conference on Computer Vision and Pattern Recognition},
  year      = {2025},
  month     = {June},
  pages     = {2609--2619},
  publisher = {IEEE},
  editor    = {IEEE Computer Society},
}

@article{yosef2023video,
  title={Video reconstruction from a single motion blurred image using learned dynamic phase coding},
  author={Yosef, Erez and Elmalem, Shay and Giryes, Raja},
  journal={Scientific Reports},
  volume={13},
  number={1},
  pages={13625},
  year={2023},
  publisher={Nature Publishing Group UK London}
}

@article{imani2024stereoscopic,
  title={Stereoscopic video deblurring transformer},
  author={Imani, Hassan and Islam, Md Baharul and Junayed, Masum Shah and Ahad, Md Atiqur Rahman},
  journal={Scientific reports},
  volume={14},
  number={1},
  pages={14342},
  year={2024},
  publisher={Nature Publishing Group UK London}
}

\end{document}